\documentclass[10pt,twocolumn,letterpaper]{article}

\usepackage[pagenumbers]{cvpr} 

\usepackage{booktabs}
\usepackage{multirow}
\usepackage{bbm}
\usepackage{amsmath}
\usepackage{placeins}
\usepackage{pifont}
\usepackage{makecell}
\usepackage{arydshln}
\usepackage{dblfloatfix}

\usepackage[skins]{tcolorbox}
\usepackage[table]{xcolor}

\definecolor{cvprblue}{rgb}{0.21,0.49,0.74}
\definecolor{lightgray}{gray}{0.8}
\definecolor{Green}{RGB}{0,150,0} 
\definecolor{Red}{named}{red}     

\newtcolorbox{myframe}[2][]{%
  left=2pt,right=2pt,enhanced,colback=white,colframe=blue,coltitle=black,
   boxrule=0.4pt,
  fonttitle=\itshape,
  attach boxed title to top left={yshift=-0.5\baselineskip-0.4pt,xshift=2mm},
  boxed title style={tile,size=minimal,left=0.1mm,right=0.1mm,
    colback=white,before upper=\strut},
  title=#2,#1
}

\usepackage[pagebackref,breaklinks,colorlinks,allcolors=cvprblue]{hyperref}
\title{MatchED: Crisp Edge Detection Using End-to-End, Matching-based Supervision}

\author{
  Bedrettin Çetinkaya \quad Sinan Kalkan\thanks{Equal senior authorship} \quad Emre Akbaş$^*$ \\
  Dept. of Computer Engineering and METU ROMER Robotics Center\\
  Ankara, Turkey\\
  {\tt\small \{bckaya, skalkan, eakbas\}@metu.edu.tr}
}

\newcommand{\CStart}[0]{\begin{center} \begin{tabular}{@{}c@{}}}
\newcommand{\CEnd}[0]{\end{tabular} \end{center}}
\newcommand{\kaynak}[1]{\hfill{\scriptsize\textcolor{gray}{#1}}}
\newcommand{\improve}[1]{\rlap{\textcolor{Green}{$^{\text{#1}}$}}}
\newcommand{\decrease}[1]{\rlap{\textcolor{Red}{$^{\text{#1}}$}}}
\newcommand{\highlight}[1]{\colorbox{gray!25}{#1}}
\newcommand{\highlightB}[1]{\colorbox{gray!15}{#1}}
\newcommand{\highlightC}[1]{\colorbox{gray!40}{#1}}
\newcommand{\highlightD}[1]{\colorbox{gray!70}{#1}}

\newcommand{\MethodLPP}{\textsc{MatchED}\xspace}
\newcommand{\Edge}{\mathbf{E}}
\newcommand{\RawEdgeR}{\mathbf{E}^r}

\newcommand{\EdgeC}{\mathbf{E}^c}
\newcommand{\GT}{\mathbf{G}}
\newcommand{\GTW}{\mathbf{\hat{G}}}

\newcommand{\TC}{\tau_c}
\newcommand{\TD}{\tau_d}

\newcommand{\cmark}{\ding{51}}
\newcommand{\xmark}{\ding{55}}

\DeclareMathOperator*{\argmin}{arg\,min}

\begin{document}
\maketitle

\begin{abstract}
Generating crisp, i.e., one-pixel-wide, edge maps remains one of the fundamental challenges in edge detection, affecting both traditional and learning-based methods. To obtain crisp edges, most existing approaches rely on two hand-crafted post-processing algorithms, Non-Maximum Suppression (NMS) and skeleton-based thinning, which are non-differentiable and hinder end-to-end optimization. Moreover, all existing crisp edge detection methods still depend on such post-processing to achieve satisfactory results.
To address this limitation, we propose \MethodLPP, a lightweight, only $\sim$21K additional parameters, and plug-and-play matching-based supervision module that can be appended to any edge detection model for joint end-to-end learning of crisp edges. At each training iteration, \MethodLPP performs one-to-one matching between predicted and ground-truth edges based on spatial distance and confidence, ensuring consistency between training and testing protocols. Extensive experiments on four popular datasets demonstrate that integrating \MethodLPP substantially improves the performance of existing edge detection models. In particular, \MethodLPP increases the Average Crispness (AC) metric by up to 2--4$\times$ compared to baseline models. Under the crispness-emphasized evaluation (CEval), \MethodLPP further boosts baseline performance by up to 20--35\% in ODS and achieves similar gains in OIS and AP, achieving SOTA performance that matches or surpasses standard post-processing for the first time.  Code is available at \url{https://cvpr26-matched.github.io}.
\end{abstract}

\section{Introduction}

\begin{figure}[t!]
\centering\footnotesize
\includegraphics[width=0.9\linewidth]{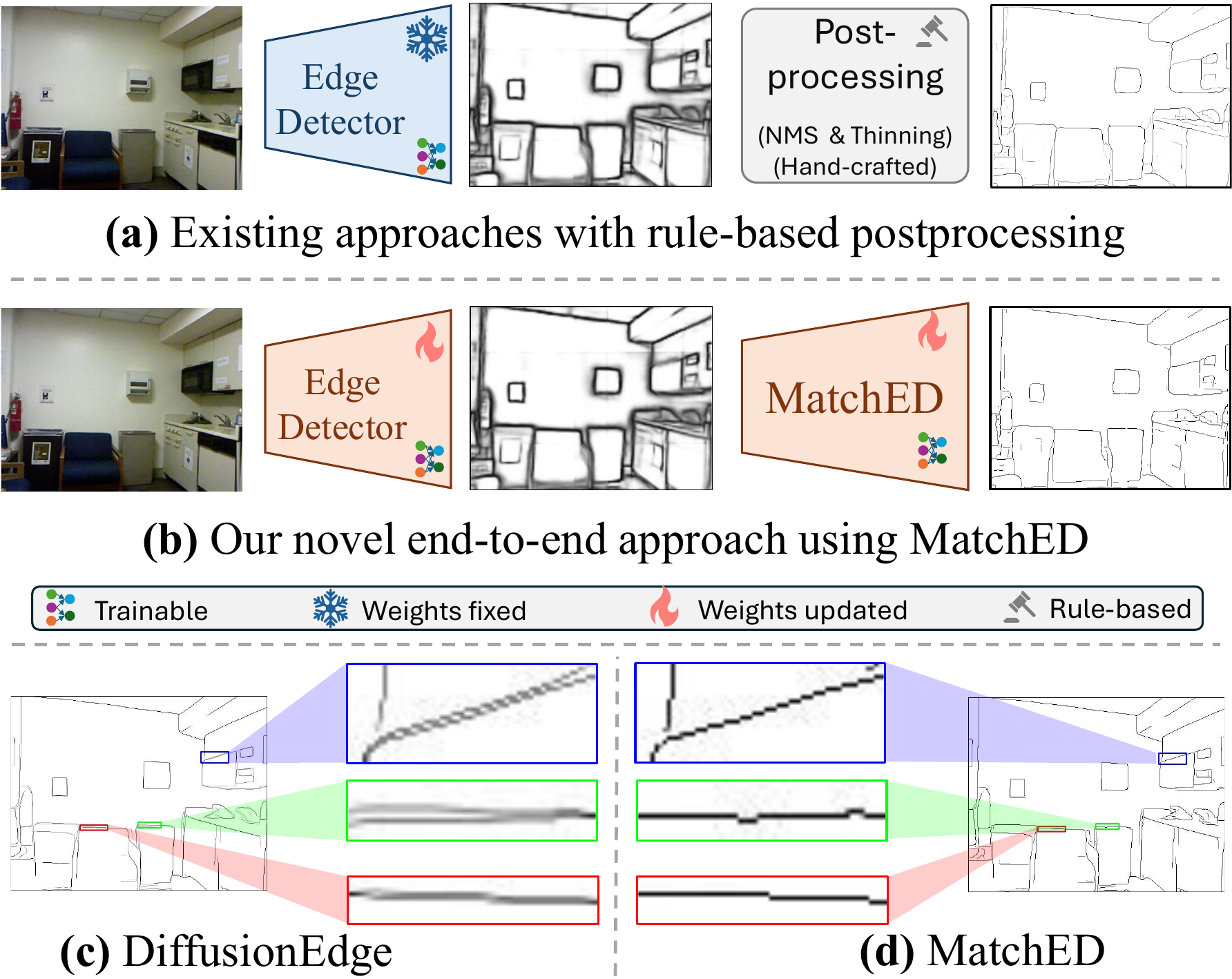}
  \caption{\textbf{(a)} Existing edge detectors require the use of a hand-crafted post-processing stage (e.g., NMS and thinning). \textbf{(b)} With \MethodLPP, we generate a one-pixel-wide edge map in an end-to-end trainable manner. The recent  DiffusionEdge's results \cite{ye2024diffusionedge} \textbf{(c)} improve when integrated with \MethodLPP \textbf{(d)}. } 
  \label{fig:teaser}
\end{figure}

Edge detection is a fundamental problem in computer vision that supports tasks such as depth estimation \cite{hu2019revisiting}, segmentation \cite{chen2016semantic,zhu2020edge,tchinda2021retinal}, and image generation \cite{zhang2023adding,zhao2023uni,wang2024omnicontrolnet} by identifying significant transitions in pixel intensity, which typically indicate object boundaries, textures, and structural features. Most research in edge detection focuses on detecting edges without explicitly accounting for their thickness. These studies \cite{canny1986computational, bertasius2015deepedge,shen2015deepcontour,xie2015holistically,maninis2016convolutional,liu2017richer,deng2018learning,he2019bi,kelm2019object,su2021pixel,pu2021rindnet,pu2022edter,zhou2023treasure,zhou2024muge,cetinkaya2024ranked,liufu2024sauge,hong2024edcssm,li2025edmb,li2025doubly} employ post-processing techniques such as non-maximum suppression (NMS) and skeleton-based thinning to produce one-pixel-wide edge maps after training (Figure \ref{fig:teaser}(a)). Despite these efforts, producing crisp edge maps remains a challenging task. 

Crispness of edge maps is crucial especially when they are used as auxiliary inputs for downstream tasks. Blurred or thick edge predictions introduce spatial uncertainty, which degrades boundary localization, ultimately worsening task performance. Precise structural cues provided by crisp edges enable models to better align semantic and geometric representations, thereby improving both the visual quality and quantitative metrics for edge-guided tasks such as image inpainting \cite{nazeri2019edgeconnect}, 3D instance segmentation \cite{roh2024edge}, and semantic segmentation \cite{yi2023edge}.

Generating crisp edges requires exact spatial matching between prediction and ground-truth \cite{ye2023delving}. Few studies \cite{deng2018learning, wang2018deep, huan2021unmixing,ye2024diffusionedge,liu2025cycle} have attempted to generate crisp edges directly; however, their performance lags behind the common ``detect and then post-process" approach, and still benefits from post-processing techniques. Producing crisp edges requires a precise spatial match between predictions and ground truth, yet imprecise labeling often causes mismatches, leading to thicker edges. Only one study \cite{ye2023delving} addresses this issue by refining labels using fixed Canny-based guidance before training, which cannot adapt to the evolving predictions during optimization and thus limits the quality of the match.

In this paper, we address the challenge of generating crisp edge maps with a novel solution, \MethodLPP (Figure \ref{fig:teaser}), which employs matching-based supervision to establish bipartite correspondences between predictions and ground truth based on distance and confidence scores at each training iteration. \MethodLPP is a lightweight, plug-and-play, learnable module that can be appended to any edge detection model for joint end-to-end training. To demonstrate its generality, we use \MethodLPP for the end-to-end training of diverse edge detectors, PiDiNet (CNN-based) \cite{su2021pixel}, RankED (transformer-based) \cite{cetinkaya2024ranked}, DiffusionEdge (diffusion-based) \cite{ye2024diffusionedge}, and SAUGE (transformer-based) \cite{liufu2024sauge}. It demonstrates consistent improvements under the crispness-emphasized evaluation (CEval), boosting state-of-the-art methods such as PiDiNet, RankED, and SAUGE by up to 20–35\% in ODS and similarly in OIS and AP. Furthermore, \MethodLPP enhances the crispness of the SOTA \textit{crisp-edge-
detection} model DiffusionEdge \cite{ye2024diffusionedge} (see Figure \ref{fig:teaser}(c–d) for visual results).  

\noindent\textbf{Contributions.} Our main contributions are as follows:
\begin{itemize}
\item We introduce a one-to-one edge matching formulation that integrates confidence scores and the evaluation distance threshold, thereby ensuring consistency between training and testing protocols.
\item \MethodLPP as a novel, trainable, plug-and-play alternative to traditional post-processing methods for producing one-pixel-wide edges. Leveraging matching-based supervision, it achieves precise edge localization and can be integrated into diverse edge detection models in an end-to-end trainable manner.
\item Extensive experiments on four popular datasets demonstrate that \MethodLPP improves the Average Crispness (AC) metric up to 2-4$\times$ compared to the original model while also surpassing standard post-processing methods in ODS, OIS, and AP across various datasets and models. 
\item We consistently outperform the SOTA models on ODS, OIS, AP, and AC using CEval on different datasets. It is the first method to match or even surpass conventional post-processing techniques (NMS + thinning) across multiple models and benchmarks.
\end{itemize}
 
\section{Related Work}
\label{sec:relWork}
\noindent \textbf{Edge Detection.} Traditional edge detection methods, such as Roberts \cite{roberts1963machine}, Canny \cite{canny1986computational}, and Sobel \cite{kittler1983accuracy}, rely on image gradients to identify edge pixels. These methods can easily yield false positive edges for pixels with high-intensity differences from neighboring regions. With advancements in deep learning, modern edge detectors \cite{bertasius2015deepedge,shen2015deepcontour,xie2015holistically,maninis2016convolutional,liu2017richer,deng2018learning,he2019bi,kelm2019object,su2021pixel,pu2021rindnet,pu2022edter,zhou2023treasure,zhou2024muge,cetinkaya2024ranked,liufu2024sauge,ye2024diffusionedge,hong2024edcssm,li2025edmb,li2025doubly} learn the complex features of the scene and produce semantically better results. Earlier CNN-based methods \cite{xie2015holistically, liu2017richer, he2019bi} use multi-scale features supervised by auxiliary outputs. Later, light-weight CNN architectures \cite{su2021pixel,soria2023tiny} are proposed for efficient edge detection. Some recent studies \cite{zhou2023treasure, cetinkaya2024ranked, zhou2024muge} leverage label uncertainty, arising from inconsistencies among annotators, to enhance the performance of edge detectors. Additionally, SAM (Segment Anything \cite{kirillov2023segment}) -based models \cite{yang2024boosting, liufu2024sauge} use features of the pre-trained SAM and give superior performances on diverse edge detection datasets. SSM (state-space models)-based models \cite{hong2024edcssm,li2025edmb} successfully adopt SSM architectures for edge detection task.

\paragraph{Standard Post-Processing in Edge Detection.} 
All existing edge detection algorithms, including simple \cite{kittler1983accuracy, roberts1963machine,jain1995machine, marr1980theory, canny1986computational} and modern ones \cite{bertasius2015deepedge,shen2015deepcontour,xie2015holistically,maninis2016convolutional,liu2017richer,deng2018learning,he2019bi,kelm2019object,su2021pixel,pu2021rindnet,pu2022edter,zhou2023treasure,zhou2024muge,cetinkaya2024ranked,liufu2024sauge,ye2024diffusionedge,hong2024edcssm,li2025edmb,li2025doubly} require post-processing to thin their raw outputs. 
This post-processing consists of two hand-crafted algorithms, applied sequentially: (i) Non-Maximum Suppression (NMS) and (ii) skeleton-based thinning \cite{guo1989parallel,lam1992thinning}. NMS preserves the strongest edge responses along the gradient direction by suppressing non-maximum pixels. However, miscalculated gradients due to noise can easily yield edges thicker than one pixel. Therefore, skeleton-based thinning iteratively refines such cases while preserving edge connectivity.

\paragraph{Crisp Edge Detection.} One of the most challenging problems in edge detection is generating a crisp edge map. Deng et al. \cite{deng2018learning} propose a new loss function, which combines cross-entropy and dice losses. Their loss function produces crisper edges compared to standard cross-entropy loss by considering the image-level similarity between prediction and ground truth. Wang et al. \cite{wang2018deep} present a refinement module to localize edge pixels precisely. Later, Huan et al. \cite{huan2021unmixing} propose a tracing loss function with context-aware fusion blocks. This loss function enables feature unmixing for improved auxiliary edge learning, while a context-aware fusion block aggregates complementary features to address side mixing. Rella et al. \cite{rella2022zero} introduce a novel vector-transform representation where each pixel predicts a unit vector pointing toward its nearest boundary. This approach mitigates the class imbalance caused by thick boundary labels and enables direct training of crisp, well-defined boundaries, particularly effective when using annotations derived from segmentation masks. Ye et al. \cite{ye2023delving} propose a Canny-guided label refinement algorithm to align ground-truth edges with RGB edges. Additionally, DiffusionEdge \cite{ye2024diffusionedge} employs a diffusion probabilistic model (DPM) in latent space by filtering latent features with a Fourier filter, achieving superior performance on various edge detection datasets based on crispness edge detection metrics. Also, CPD \cite{liu2025cycle} presents a new architecture for efficient crisp edge detection using both edge prior and multi-scale information. Although all of these studies aim to predict crisp edge maps, they still benefit from
post-processing methods (NMS + skeleton-based thinning) used in standard edge detection evaluation (SEval), unlike our work.

\begin{table}[hbt!]
    
    \caption{Comparative summary of crisp edge detection methods.}
    \centering
    \resizebox{0.75\columnwidth}{!}{%
    \begin{tabular}{lcc}
        \toprule
        \multirow{2}{*}{Method} & Integrable to & Post-process \\ 
                                & any model?    & Free? \\
        \midrule
        LPCB \cite{deng2018learning}         & \cmark & \xmark \\
        CED \cite{wang2018deep}              & \xmark & \xmark \\
        CATS \cite{huan2021unmixing}         & \cmark & \xmark \\
        GLR \cite{ye2023delving}             & \cmark & \xmark \\
        DiffusionEdge \cite{ye2024diffusionedge} & \xmark & \xmark \\
        CPD \cite{liu2025cycle}              & \xmark & \xmark \\
        \midrule
        \MethodLPP~ (\textbf{Ours})          & \cmark & \cmark \\
        \bottomrule
    \end{tabular}
    }
    \label{tab:comparative_summary}
\end{table}

\noindent \textbf{Comparative Summary.} As summarized in Table~\ref{tab:comparative_summary}, all prior methods show performance gains when hand-crafted post-processing is applied to their outputs, indicating that their raw predictions lack crispness. Moreover, only a few existing studies can be integrated with any edge detection model. \MethodLPP, in contrast, is a plug-and-play method that provides post-processing-free crisp edges across a diverse set of edge detectors.

\section{Methodology: \MethodLPP}

\begin{figure*}
\centering
\includegraphics[width=0.8\linewidth]{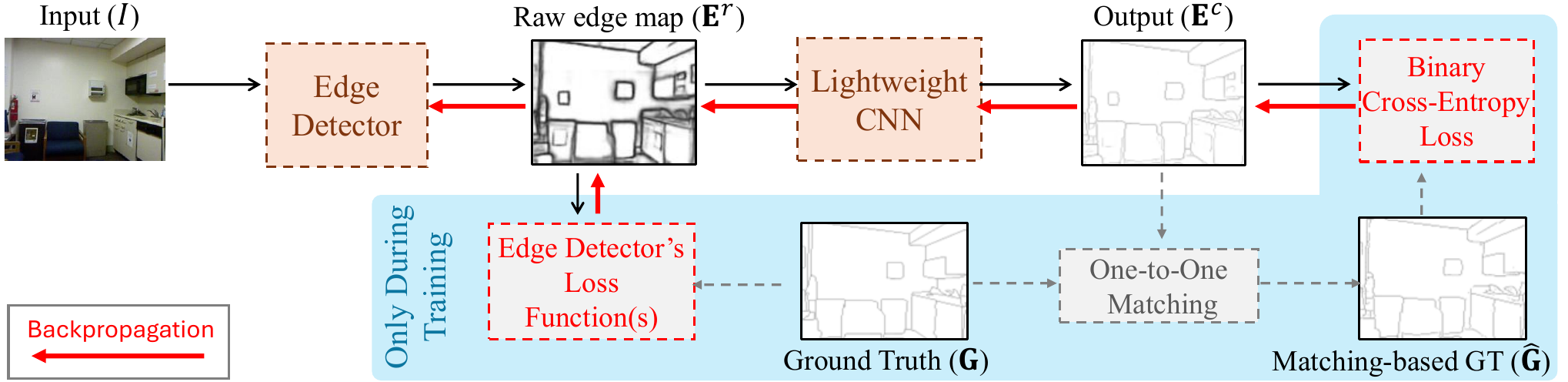}
  \caption{An overview of \MethodLPP, which can be easily integrated into any edge detector pipeline. A lightweight CNN consists of five blocks, each containing Conv2D, ReLU, and normalization layers, followed by a final Conv2D layer with a sigmoid activation. Therefore, \MethodLPP adds only approximately 21K parameters.}
  
  \label{fig:arch}
\end{figure*}

We introduce \MethodLPP (Figure \ref{fig:arch}), a light-weight plug-and-play module that can be integrated into any deep edge detection model to generate crisp edge maps. Achieving such crispness requires precise spatial alignment between the predicted and ground-truth edges, which is often hindered by annotation noise and inherent localization ambiguities. We provide a more detailed analysis of this thickness in the supplementary material.  Prior works typically tolerate these misalignments through hand-crafted post-processing or spatially relaxed supervision strategies, which often result in thick edge predictions to compensate for non-aligned annotations, or by refining labels before training \cite{ye2023delving}, which cannot achieve optimal alignment between the model predictions and the ground truth due to a fixed Canny guidance. To overcome these limitations, \MethodLPP is trained end-to-end with a novel matching-based supervision signal that explicitly aligns predicted and ground-truth edges during training. The module is trained end-to-end using a novel supervision signal derived from matching the predicted and ground-truth edges at each training iteration.
 
\subsection{Problem Definition and Notation}

Edge detection is the problem of finding pixels in an $W \times H$ image $I\in \mathbb{R}^{W \times H \times 3}$ that belong to semantically relevant intensity changes as an edge map $\Edge\in \mathbb{R}^{W \times H}$. A common challenge in edge detection is the thickness of the estimated edges, which is often addressed by using a hand-crafted post-processing stage. 

Our goal in this study is to design a differentiable module, \MethodLPP, that can be integrated into any edge detection model producing a raw edge map $\RawEdgeR \in \mathbb{R}^{W \times H}$, to generate a crisp edge map $\EdgeC \in \mathbb{R}^{W \times H}$ in an end-to-end trainable manner.

\subsection{Integrating \MethodLPP into Existing Detectors} Edge detection model $f$ parametrized by $\theta_r$ takes an RGB image $I$ as an input and produces a raw  edge map $\RawEdgeR$:
\begin{equation}
\RawEdgeR = f(I; \theta_r).
\label{eq:edge-detector}
\end{equation}

To combine \MethodLPP~with $f$, we append it to the last stage of $f$; i.e. the output of $f$ is the input of \MethodLPP:
\begin{equation}
 \EdgeC = \MethodLPP\left(\RawEdgeR; \theta_c\right),
 \label{eq:matched}
\end{equation}
where $\theta_c$ represents the parameters of \MethodLPP.

\subsection{Alignment \& Matching with Ground Truth}
\noindent\textbf{Spatial Alignment Cost.} In edge detection datasets, positions of edges in ground truth ($\GT$) may not perfectly align with the estimated edges in $\EdgeC$ due to crude manual labeling. In the edge detection literature \cite{su2021pixel,zhou2023treasure,cetinkaya2024ranked,liufu2024sauge}, such localization errors are tolerated during evaluation by applying a dataset-specific distance threshold. To maintain consistency between training and evaluation, we introduce a one-to-one matching formulation between the ground-truth edge map $\GT$ and the predicted crisp edge map $\EdgeC$, which jointly considers edge confidence scores and spatial distance constraints to quantify their alignment through a confidence-weighted matching cost.

For each pixel in the predicted map $\EdgeC$, denoted as $\mathbf{p_c}$, and each pixel in the ground truth $\GT$, denoted as $\mathbf{p_g}$, we compute the cost matrix $C$ as follows:
\begin{equation}\footnotesize
\label{eq:costMat}
C(\mathbf{p_c}, \mathbf{p_g}) =
\begin{cases}
d(\mathbf{p_c}, \mathbf{p_g}) - \alpha \EdgeC(\mathbf{p_c}), &
\text{if } 
\left(
\vcenter{
  \hbox{
    $\begin{aligned}
      &\EdgeC(\mathbf{p_c}) \ge \TC,\\
      &\GT(\mathbf{p_g}) = 1,\\
      &d(\mathbf{p_c}, \mathbf{p_g}) < \TD
    \end{aligned}$
  }
}
\right)
\\
\infty, & \text{otherwise.}
\end{cases}
\end{equation}

\noindent Here, $\mathbf{p_c}$ and $\mathbf{p_g}$ represent 2D coordinates of pixels in $\EdgeC$ and $\GT$, respectively. The term $d(\mathbf{p_c}, \mathbf{p_g})$ denotes the Manhattan distance between these pixels, which captures their spatial distance.  $\EdgeC(\mathbf{p_c})$ indicates the confidence score of pixel $\mathbf{p_c}$, and $\alpha$ is a weighting factor that balances the effect of this confidence in the overall matching cost.

The first condition, $(\EdgeC(\mathbf{p_c}) \ge \TC)$, filters out low-confidence predictions in $\EdgeC$   using a confidence threshold $\TC$. The second condition, $(\GT(\mathbf{p_g}) = 1)$, ensures that only ground-truth edge pixels are considered as valid matches. The final condition, $(d(\mathbf{p_c}, \mathbf{p_g}) < \TD)$, limits the matching process to a local neighborhood around each predicted edge pixel, controlled by the distance threshold $\TD$.

By integrating these conditions into a single formulation, the cost matrix $C$ effectively encodes both spatial and confidence relations. A higher-confidence prediction $\EdgeC(\mathbf{p_c})$ reduces the corresponding matching cost, encouraging alignment with nearby ground-truth edge pixels. Conversely, pixels that do not satisfy the above conditions are assigned an infinite cost, preventing invalid or distant matches. This \emph{confidence-weighted} and \emph{locality-aware} formulation allows our method to establish pixel-level correspondences between predicted and ground-truth edge maps.

\noindent\textbf{Optimal 1-to-1 Matching with Ground Truth.} To produce crisp edges, edges in $\EdgeC$ should perfectly align with edges in $\GT$. Otherwise, the edge detection model generates responses not only for a single pixel but also for its neighboring regions, leading to thicker edges. To alleviate this problem, each response should be uniquely assigned to a ground-truth edge. To achieve this, we formulate the problem as an optimal bipartite matching between $\EdgeC$ and $\GT$:
\begin{equation}
\label{eq:BiPartite}
\sigma^*=\argmin_{\sigma \in S_{n}} \sum_{i}^n C(i, \sigma(i)),
\end{equation}
where $n= W \times H$; $S_{n}$ is the set of all permutations; $C(i, \sigma(i))$ represents the cost of assigning row $i$ to column $\sigma(i)$; and, $\sigma^*$ (from Equation \ref{eq:costMat}) is the optimal assignment. We compute $\sigma^*$ using linear sum assignment following prior work in object detection \cite{stewart2016end,carion2020end}. Using $\sigma^*$, we get a matching-based ground-truth label $\GTW$:
\begin{equation}
\label{eq:newG}
\GTW(i)=\GT(\sigma^*(i)),
\end{equation}
which perfectly matches pixels in $\EdgeC$ with edge pixels in $\GT$, that satisfy conditions specified in Eq. \ref{eq:costMat}. However, some edge pixels in $\GT$ may remain unmatched due to the distance threshold $\TD$, i.e., when no predicted response exists within $\TD$ of an edge pixel $p_g$ in $\GT$. To recover these pixels, we directly assign them to $\GTW$ as:
\begin{equation}
\label{eq:unmatched}
\GTW(\mathbf{p_g}) = 1, \quad \text{if } \mathbf{p_g} \text{ is unmatched and } \GT(\mathbf{p_g}) \text{ = 1.}
\end{equation}
This strategy ensures that regions around unmatched ground-truth pixels remain active, enabling their potential inclusion in the matching in subsequent iterations. Details on the time complexity of our matching algorithm are presented in the supplementary material.

\subsection{Training Objective} The parameters $\theta_r$ and $\theta_c$ are jointly optimized. $\theta_r$ is optimized using ground-truth $\GT$ and the loss functions of $f$, which depend on the corresponding model $f$ and can include any loss function, such as weighted binary cross-entropy loss \cite{xie2015holistically,liu2017richer}. 
%
%
%
%

We use binary cross-entropy loss between $\EdgeC$ and $\GTW$ to learn $\theta_c$:
\begin{equation}
\label{eq:bce}\footnotesize
\mathcal{L}_{\MethodLPP} = - \sum_{i=1}^{N} \left[ \GTW_i \log(\EdgeC) + (1 - \GTW_i) \log(1 - \EdgeC)\right],
\end{equation}
where $N$ is the number of different ground truth annotations. Hence, our total loss is the weighted combination of loss of \MethodLPP and model $f$:
\begin{equation}
  \mathcal{L}_\text{total} = \beta\mathcal{L}_{\MethodLPP} + \mathcal{L}_{f}, 
\end{equation} where $\beta$ controls the contribution of \MethodLPP.

\noindent\textbf{Model.} We model \MethodLPP (Eq. \eqref{eq:matched}) using a lightweight CNN with only $\sim$21K parameters, consisting of five standard convolution blocks. Model details can be found in the supplementary material. 
\section{Experiments}
\label{sec:experiments}
\subsection{Experimental Setup and Details}
\textbf{Datasets.} We used the following four popular edge detection datasets to evaluate our method:

\underline{BSDS500 \cite{arbelaez2010contour}} is a standard benchmark for contour and edge detection, containing 500 natural images (481$\times$321 pixels) split into 200 training, 100 validation, and 200 test images. Each image has multiple human-annotated ground-truth edge maps.

\underline{NYUD-v2 \cite{silberman2012indoor}} contains 1,449 indoor scenes with RGB and depth images (576$\times$448). Edge detection ground-truth is derived from object segmentation masks, using object boundaries as edge annotations. The dataset is split into 381 training, 414 validation, and 654 test images.

\underline{BIPED \cite{soria2023dense}} is an edge detection dataset of 250 outdoor images (1280$\times$720), with 200 training and 50 test images. Each image has a single ground-truth label annotated by computer vision professionals. We used the updated official version of the dataset.

\underline{Multi-cue \cite{mely2016systematic}} consists of 100 images (1280$\times$720 resolution), each annotated with six edge labels and five boundary labels. Due to the absence of an official split, we used the proposed allocation for the training and test sets for each integrated model.

\noindent \textbf{Base Models.}
To demonstrate the generalization capability of \MethodLPP, we select four SOTA models with different architectures and loss functions, on top of which we integrate our \MethodLPP: (i) PiDiNet \cite{su2021pixel}, a CNN-based model optimized with weighted binary cross-entropy, (ii) RankED \cite{cetinkaya2024ranked}, a transformer-based model optimized with ranking-based losses, (iii) DiffusionEdge \cite{ye2024diffusionedge}, a diffusion-based model optimized with a combination of multiple loss functions, and (iv) SAUGE \cite{liufu2024sauge}, a SAM-based model optimized with a combination of multiple losses.

\noindent \textbf{Implementation and Training Details.} 
For each base model, we used the officially published code and its original settings; we did not perform any hyperparameter  (learning rate, batch size, optimizer, etc.) tuning. 

We train only the base edge detector $f$ during the first $N/2$ epochs, then jointly train $f$ and \MethodLPP for the remaining $N/2$ epochs, where $N$ is the total number of epochs for $f$ in its official paper. This training strategy is adopted because \MethodLPP can generate accurate and crisp edges only if $f$ produces reliable raw edge maps.

Hyperparameter settings for each integrated model and dataset are provided in the supplementary material. Additionally, we use a distance threshold parameter $\TD$, which represents the maximum allowable distance between predicted and ground-truth edge maps during the matching cost calculation in Eq. \ref{eq:costMat}. The values of $\TD$ are set to 8, 4, 11, and 11 for the NYUD-v2, BSDS, Multi-Cue, and BIPED datasets, respectively, and are applied consistently across all base models. These values align with the evaluation protocol used in standard edge detection benchmarks. Additionally, we apply a $5 \times 5$ box blur kernel to the outputs of RankED \cite{cetinkaya2024ranked} and SAUGE \cite{liufu2024sauge} to reduce grid-like artifacts \cite{yang2024denoising} in the raw edge maps due to their transformer-based architectures. This trick is not used for other base models.

\noindent \textbf{Performance Measures.}
To evaluate the performance of edge detection models, we use four metrics: (i) Optimal Dataset Scale (ODS), which measures the best F-score using a single threshold applied across the entire dataset; (ii) Optimal Image Scale (OIS), which computes the best F-score for each image individually and then averages them; (iii) Average Precision (AP), which represents the area under the precision-recall curve, reflecting overall detection quality across thresholds; and (iv) Average Crispness (AC) \cite{ye2023delving}, which quantifies crispness of raw predicted edges.

Moreover, we report these metrics under two evaluation protocols following previous works \cite{ye2024diffusionedge,huan2021unmixing,ye2023delving}: (i) Standard Evaluation (SEval), which applies Non-Maximum Suppression (NMS) and skeleton-based thinning to produce a one-pixel-wide edge map, and (ii) Crispness-Emphasized Evaluation (CEval), which uses the raw predicted edge map without post-processing.

\subsection{Exp 1: Comparison w/ Crisp Edge Detectors}

First, we evaluate the crispness of \MethodLPP and compare its performance with other crisp edge detectors, LPCB \cite{deng2018learning}, CATS \cite{huan2021unmixing}, GLT \cite{ye2023delving}, and DiffusionEdge \cite{ye2024diffusionedge}. As shown in Table \ref{tab:ch5_SOTA_crispEdge}, our method produces significantly crisper edges than other approaches across all datasets. It outperforms the second-best method (DiffusionEdge \cite{ye2024diffusionedge}) by +0.454, +0.348, and +0.092 on the BSDS, Multi-Cue, and BIPED datasets, respectively.

To the best of our knowledge, existing crisp edge detection methods fall short of matching the performance of standard post-processing techniques. In contrast, our method achieves comparable results, and in some cases even outperforms them as will be shown in Section \ref{sect:SOTAcomp_ch5}, without relying on such post-processing. 

\begin{myframe}{Main Result}
Experiment 1 shows that  \MethodLPP consistently outperforms SOTA crisp edge detection models by a large margin. In terms of the AC metric, \MethodLPP provides up to 1.7–4$\times$ gains on diverse datasets. 
\end{myframe}

\begin{table}[t]
\caption{\textbf{Exp 1:} Comparison of SOTA crisp edge detection methods and ours in terms of Average Crispness (AC) metric. Bold and underlined numbers indicate the best and second-best results, respectively. Results are taken from \cite{ye2024diffusionedge}. D: Dice Loss \cite{deng2018learning}, T: Tracing-Loss \cite{huan2021unmixing}, R: Refined Label \cite{ye2023delving}.}
\label{tab:ch5_SOTA_crispEdge}
\centering
\resizebox{0.9\columnwidth}{!}{%
\begin{tabular}{llccc}
\toprule
\multicolumn{2}{l}{\textbf{Method}} & \textbf{BSDS} & \textbf{Multi-cue} & \textbf{BIPED} \\
\midrule
PiDiNet $+$ D \cite{deng2018learning} & \kaynak{(ECCV18)} & .306 & .208 & .340 \\
PiDiNet $+$ T \cite{huan2021unmixing} & \kaynak{(PAMI22)}  & .333 & .217 & .296 \\
PiDiNet $+$ R \cite{ye2023delving} & \kaynak{(TIP23)}     & .424 & .424 & .512 \\
DiffEdge \cite{ye2024diffusionedge} & \kaynak{(AAAI24)} & \underline{.476} & \underline{.462} & \underline{.849} \\
\midrule
\multicolumn{2}{l}{PiDiNet $+$ \MethodLPP} & \textbf{.930} & \textbf{.810} & \textbf{.941} \\
\bottomrule
\end{tabular}
}
\end{table}

\subsection{Exp 2: Comparison w/ All Edge Detectors}
\label{sect:SOTAcomp_ch5}
This section compares \MethodLPP with the state-of-the-art (SOTA) methods on NYUD-v2, BSDS, and BIPED-v2 datasets using both crispness-emphasized evaluation (CEval) and standard evaluation (SEval). Also, results on Multi-Cue dataset can be found in the supplementary material.

\label{sect:ch5_BSDS}\noindent\textbf{BSDS.} \Cref{tab:ch5_SOTA_crisp_BSDS} presents comparison with SOTA approaches using CEval and SEval protocols. 
Integrating \MethodLPP into PiDiNet \cite{su2021pixel} yields gains of +0.222, +0.224, and +0.717 in ODS, OIS, and AC, respectively. For RankED \cite{cetinkaya2024ranked}, improvements are +0.188 (ODS), +0.192 (OIS), +0.179 (AP), and +0.438 (AC), while integration into SAUGE \cite{liufu2024sauge} gives +0.156 (ODS), +0.154 (OIS), +0.144 (AP), and +0.631 (AC). Also, DiffusionEdge-based \MethodLPP~ yields +0.084 (ODS), +0.094 (OIS), +0.144 (AP), and +0.474 (AC).

To further demonstrate the efficiency of \MethodLPP, we compare its CEval results with the SEval performance of SOTA methods. PiDiNet $+$ \MethodLPP adds +0.011 (ODS) and +0.008 (OIS), and SAUGE gains +0.008 (ODS), +0.002 (OIS), and +0.013 (AP). RankED shows slight reductions (-0.011 ODS, -0.005 OIS, -0.03 AP) due to down-sampled predictions: its edge maps are produced at 0.25× scale and upsampled to input resolution before computing the loss, introducing interpolation artifacts that slightly affect its performance. Note that SAUGE $+$ \MethodLPP achieves the best performance across all metrics among SOTA methods that rely on post-processing.

\begin{table}[ht]
\caption{\textbf{Exp 2:} Comparison of SOTA results under Standard (SEval) and Crispness-emphasized (CEval) evaluations on the BSDS dataset. 
The best and second-best results are shown in bold and underlined, respectively. \textcolor{Green}{+Green} highlights improvements compared to the base model.}
\label{tab:ch5_SOTA_crisp_BSDS}
\centering
\setlength{\tabcolsep}{2pt}
\resizebox{0.47\textwidth}{!}{%
\begin{tabular}{l@{\hspace{0pt}}lcccccccc}
\toprule
\multicolumn{2}{c}{} & \multicolumn{3}{c}{\textbf{CEval}} & & \multicolumn{3}{c}{\textbf{SEval}} \\
\cmidrule(lr){3-5} \cmidrule(lr){7-9}
\multicolumn{2}{c}{\textbf{Method}} & \textbf{ODS\ \ } & \textbf{\ \ \ OIS\ \ } & \textbf{\ \ AP\ \ \ } & \textbf{\ \ \ AC\ \ } & \textbf{\ ODS} & \textbf{OIS} & \textbf{AP} \\
\midrule
HED   \cite{xie2015holistically}      & \kaynak{(ICCV15)}  & .588 & .608 & --   & .215 & .788 & .808 & --   \\
RCF   \cite{liu2017richer}            & \kaynak{(CVPR17)}  & .585 & .604 & --   & .189 & .798 & .815 & --   \\
LPCB  \cite{deng2018learning}         & \kaynak{(CVPR18)}  & .693 & .700 & --   & --   & .800 & .816 & --   \\
BDCN  \cite{he2019bi}                 & \kaynak{(CVPR19)}  & .636 & .650 & .643 & .233 & .817 & .833 & .883 \\
\highlightB{PiDiNet} \cite{su2021pixel}            & \kaynak{(ICCV21)}  & .578 & .587 & --   & .202 & .789 & .803 & --   \\
EDTER \cite{pu2022edter}              & \kaynak{(CVPR22)}  & .698 & .706 & --   & .288 & .824 & .841 & .880 \\
UAED  \cite{zhou2023treasure}         & \kaynak{(CVPR23)}  & .722 & .731 & --   & .227 & .829 & .847 & .892 \\
MuGE  \cite{zhou2024muge}             & \kaynak{(CVPR24)}  & .720 & .728 & .800 & .299 & .831 & .847 & .886 \\
\highlight{DiffEdge} \cite{ye2024diffusionedge} & \kaynak{(AAAI24)} & .749 & .754 & --   & .476 & .834 & .848 & --   \\
\highlightC{RankED} \cite{cetinkaya2024ranked}     & \kaynak{(CVPR24)}  & .631 & .637 & .686 & .213 & .824 & .840 & \underline{.895} \\
EdgeSAM \cite{yang2024boosting}       & \kaynak{(TII24)}   & --   & --   & --   & --   & .838 & .852 & .893 \\
EDMB \cite{li2025edmb}                & \kaynak{(WACV25)}  & --   & --   & --   & --   & .837 & .851 & --   \\
\highlightD{SAUGE} \cite{liufu2024sauge}           & \kaynak{(AAAI25)}  & .700 & .714 & .768 & .215 & \textbf{.847} & \textbf{.868} & \textbf{.898} \\
\midrule
\multicolumn{2}{l}{\highlightB{PiDiNet} $+$ \MethodLPP}          & .800\improve{+.2} & .811\improve{+.2} & .866 & \underline{.919}\improve{+.7} & .798 & .800 & .859 \\
\multicolumn{2}{l}{\highlight{DiffEdge} $+$ \MethodLPP}    & \underline{.833}\improve{+.1} & \underline{.848}\improve{+.1}   & \underline{.895}   & \textbf{.950}\improve{+.5} & .833   & 848  & .885      \\
\multicolumn{2}{l}{\highlightC{RankED} $+$ \MethodLPP}          & .819\improve{+.2} & .829\improve{+.2} & .865\improve{+.2} & .651\improve{+.4} & .812 & .830 & .862 \\
\multicolumn{2}{l}{\highlightD{SAUGE} $+$ \MethodLPP} & \textbf{.854}\improve{+.2} & \textbf{.870}\improve{+.2} & \textbf{.912}\improve{+.1} & .852\improve{+.6} & \underline{.846} & \underline{.864} & .891 \\
\bottomrule
\end{tabular}
}
\end{table}

\noindent\textbf{NYUD-v2.}
We integrate \MethodLPP into three models: (i) PiDiNet \cite{su2021pixel}, (ii) RankED \cite{cetinkaya2024ranked}, and (iii) DiffusionEdge \cite{ye2024diffusionedge}, and compare their performance with SOTA methods (Table \ref{tab:ch5_SOTA_crisp_NYUD-RGB}). \MethodLPP significantly boosts performance across all base models under CEval. For example, it improves RankED by +0.298, +0.289, +0.22, and +0.74 percentage points in ODS, OIS, AP, and AC, respectively, and PiDiNet by +0.337, +0.325, and +0.757 in ODS, OIS, and AC. Even DiffusionEdge, which already produces crisp edges, gains +0.023, +0.025, and +0.091 in ODS, OIS, and AC, confirming \MethodLPP’s generalizability and effectiveness.

Applying standard post-processing to \MethodLPP outputs does not improve performance and can reduce performance (e.g., DiffusionEdge-based ODS, OIS, AP in SEval), indicating the crispness of the predicted edge maps. SEval results show \MethodLPP performs comparably to standard post-processing, with increases for PiDiNet (+0.003 ODS, +0.002 OIS) and slight decreases for RankED (-0.005 ODS, -0.009 OIS, -0.02 AP) and the officially published results of DiffusionEdge (-0.002 ODS, -0.004 OIS). However, compared to our training of DiffusionEdge, \MethodLPP improves ODS by +0.003 while AP drops by -0.02; OIS remains unchanged.

We also evaluate the zero-shot performance of \MethodLPP, following SAUGE \cite{liufu2024sauge}, by training on BSDS and testing on NYUD-v2. Despite being unable to fully reproduce SAUGE’s reported results using the officially published model, \MethodLPP achieves performance nearly identical to standard post-processing, except for a -0.003 drop in OIS according to our evaluation.

\begin{table}[t]
\caption{\textbf{Exp 2:} Comparison of SOTA results under Standard (SEval) and Crispness-emphasized (CEval) evaluations on the NYUD-v2 dataset using RGB images. ODS: Optimal Dataset Score, OIS: Optimal Image Score, AP: Average Precision, AC: Average Crispness. The best and second-best results for supervised methods (upper part of the table) are shown in bold and underlined, respectively. The bottom section presents zero-shot results, where training is conducted on BSDS and evaluation is performed on NYUD. * denotes results from our training and $\dag$ denotes results obtained using the official checkpoint. \textcolor{Green}{+Green} highlights improvements compared to the base model.}
\label{tab:ch5_SOTA_crisp_NYUD-RGB}
\centering
\setlength{\tabcolsep}{2pt}

\resizebox{0.475\textwidth}{!}{%
\begin{tabular}{l@{\hspace{0pt}}lcccccccc}
\toprule
\multicolumn{2}{c}{} & \multicolumn{3}{c}{\textbf{CEval}} & & \multicolumn{3}{c}{\textbf{SEval}} \\
\cmidrule(lr){3-5} \cmidrule(lr){7-9}
\multicolumn{2}{c}{\textbf{Method}} & \textbf{ODS\ \ } & \textbf{\ \ \ OIS\ \ } & \textbf{\ \ AP\ \ \ } & \textbf{\ \ \ AC\ \ } & \textbf{\ ODS} & \textbf{OIS} & \textbf{AP} \\

\midrule
HED   \cite{xie2015holistically}          & \kaynak{(ICCV15)}  & .387 & .404 & --   & --   & .722 & .737 & --   \\
RCF   \cite{liu2017richer}                & \kaynak{(CVPR17)}  & .398 & .413 & --   & --   & .745 & .759 & --   \\
LPCB  \cite{deng2018learning}             & \kaynak{(CVPR18)}  & --   & --   & --   & --   & .739 & .754 & --   \\
BDCN  \cite{he2019bi}                     & \kaynak{(CVPR19)}  & .426 & .450 & .362 & .162 & .748 & .762 & .778 \\
\highlight{PiDiNet} \cite{su2021pixel}                & \kaynak{(ICCV21)}  & .399 & .424 & --   & .173 & .733 & .747 & --   \\
EDTER \cite{pu2022edter}                  & \kaynak{(CVPR22)}  & .430 & .457 & --   & .195 & .774 & .789 & .797 \\
Diff.Edge \cite{ye2024diffusionedge}  & \kaynak{(AAAI24)}  & .732 & .738 & --   & .846 & .761 & .766 & --   \\
\highlightC{Diff.Edge*} \cite{ye2024diffusionedge} & \kaynak{(AAAI24)}  & .727 & .733 & .693 & .863 & .755 & .762 & .723 \\
\highlightD{RankED} \cite{cetinkaya2024ranked}         & \kaynak{(CVPR24)}  & .477 & .490 & .483 & .146 & .780 & \underline{.793} & \textbf{.826} \\
EdgeSAM \cite{yang2024boosting}           & \kaynak{(TII24)}   & --   & --   & --   & --   & \textbf{.783} & \textbf{.797} & .805 \\
EDMB \cite{li2025edmb}                    & \kaynak{(WACV25)}  & --   & --   & --   & --   & \textbf{.783} & .789 & -- \\
\midrule
\multicolumn{2}{l}{\highlight{PiDiNet} $+$ \MethodLPP}  & .736\improve{+.3} & .749\improve{+.3} & \underline{.777} & \underline{.930}\improve{+.7} & .736 & .749 & .756 \\
\multicolumn{2}{l}{\highlightC{Diff.Edge} $+$ \MethodLPP} & \underline{.759}\improve{+.03} & \underline{.762}\improve{+.03} & .703\improve{+.01} & \textbf{.937}\improve{+.07} & .758 & .762 & .703 \\
\multicolumn{2}{l}{\highlightD{RankED} $+$ \MethodLPP}       & \textbf{.775}\improve{+.3} & \textbf{.784}\improve{+.3} & \textbf{.806}\improve{+.3} & .886\improve{+.7} & .774 & .784 & .802 \\
\midrule
\multicolumn{9}{c}{\textit{Zero-shot Performance (trained on BSDS, evaluated on NYUD)}} \\
\midrule
SAUGE \cite{liufu2024sauge}             & \kaynak{(AAAI25)}  & --   & --   & --   & --   & .794 & .803 & .813 \\
\highlightB{SAUGE}\dag \cite{liufu2024sauge}            & \kaynak{(AAAI25)}  & .491 & .517 & .554 & .187 & .758 & .782 & .810 \\
\multicolumn{2}{l}{\highlightB{SAUGE} $+$ \MethodLPP}    & .758\improve{+.2} & .779\improve{+.2} & .813\improve{+.2} & .852\improve{+.6} & .760 & .784 & .811 \\
\bottomrule
\end{tabular}
}
\end{table}

\noindent\textbf{BIPED-v2.} We also evaluate the performance of our models on the BIPED-v2 dataset, which is designed for edge detection rather than contour or boundary detection, as shown in Table \ref{tab:ch5_SOTA_crisp_BIPED}. While SOTA methods are evaluated after post-processing, our method is evaluated without post-processing. PiDiNet $+$ \MethodLPP has the best performance for all metrics among SOTA methods. Additionally, it improves PiDiNet \cite{su2021pixel} performance by +0.008, +0.008, and +0.02 in ODS, OIS, and AP metrics, respectively.

\begin{myframe}{Main Result}
Experiment 2 shows that  \MethodLPP consistently improves performance, particularly under the crispness-emphasized (CEval) protocol, with substantial gains in ODS, OIS, AP, and AC metrics. Additionally, \MethodLPP generates inherently crisp edge maps, often surpassing or matching standard post-processing, highlighting its generalizability and effectiveness across diverse datasets.
\end{myframe}

\begin{table}
\caption{\textbf{Exp 2:} Comparison of base models and base models with \MethodLPP on BIPED-v2 dataset. \MethodLPP results are reported without post-processing, while base model results include
post-processing. 
Best and second-best results are shown in bold and underlined, respectively. * denotes results of our training.}
\label{tab:ch5_SOTA_crisp_BIPED}
\centering
\resizebox{0.77\columnwidth}{!}{%
\begin{tabular}{l@{\hspace{0pt}}lc@{\hspace{20pt}}c@{\hspace{20pt}}c}
\toprule
\multicolumn{2}{c}{\textbf{Method}} & \textbf{ODS} & \textbf{OIS} & \textbf{AP} \\
\midrule
\rowcolor{lightgray} RCF \cite{liu2017richer}               & \kaynak{(CVPR17)} & .849 & .861 & .906 \\
BDCN \cite{he2019bi}                   & \kaynak{(CVPR19)} & .890 & .899 & .934 \\
CATS \cite{huan2021unmixing}           & \kaynak{(PAMI21)} & .887 & .892 & .817 \\
\highlight{PiDiNet} \cite{su2021pixel}             & \kaynak{(ICCV21)} & .892 & .897 & \underline{.951} \\
DexiNed \cite{soria2023dense}          & \kaynak{(Pat.Rec.22)} & \underline{.895} & \underline{.900} & .927 \\
\highlightD{DiffEdge*} \cite{ye2024diffusionedge} & \kaynak{(AAAI24)} & .885 & .886 & .905 \\
\midrule
\multicolumn{2}{l}{\highlight{PiDiNet} $+$ \MethodLPP}   & \textbf{.900}\improve{+.008} & \textbf{.905}\improve{+.008} & \textbf{.971}\improve{+.02} \\
\multicolumn{2}{l}{\highlightD{Diffusion} $+$ \MethodLPP}  & .889\improve{+.004} & .892\improve{+.006} & .903\decrease{-.002} \\
\bottomrule
\end{tabular}
}
\end{table}

\subsection{Exp 3: Ablation Analysis}
\noindent{\textbf{Effects of hyperparameters.}} We analyze the impact of  confidence threshold $\TC$ (Eq.~\ref{eq:costMat}), distance threshold $\TD$ (Eq.~\ref{eq:costMat}), and confidence score weight $\alpha$ (Eq.~\ref{eq:costMat}) on the performance of \MethodLPP, using PiDiNet \cite{su2021pixel} on BSDS dataset. The results are summarized in Table \ref{tab:abl_exp}(a, b, c) for each hyperparameter, respectively. Overall, \MethodLPP demonstrates robustness across reasonable ranges of these hyperparameters. A more detailed discussion 
is provided in the supplementary material.

\noindent\textbf{Number of hyperparameters in \MethodLPP.}
We compare \MethodLPP~with standard post-processing in terms of the number of hyperparameters.
While NMS uses three hyperparameters, $r$ (window size for suppression check), $s$ (border fade size), and $e$ (edge magnitude multiplier), and skeleton-based thinning has no parameters, our proposed method (\MethodLPP) introduces four hyperparameters: $\TC$, $\TD$, $\alpha$, and $\beta$. Notably, $\TD$ does not require additional search, as it is identical to the maximum distance tolerance already employed in the evaluation procedure of edge detection results. Therefore, the number of hyperparameters is the same for both approaches.

\begin{myframe}{Main Result}
Experiment 3 shows that \MethodLPP provides reliable performance across a wide range of hyperparameter values. \MethodLPP effectively uses three hyperparameters, comparable to NMS, while outperforming standard post-processing as shown in Experiment 2.
\end{myframe}

\subsection{Exp 4: Efficiency Analysis}

\noindent{\textbf{Parameter overhead.}} We analyze how \MethodLPP~increases the overall parameter count when integrated into existing models. 
\MethodLPP~introduces only a minimal additional cost, $\sim$21K  parameters. Even when integrated into the lightweight model PiDiNet~\cite{su2021pixel}, it increases the number of parameters by just 3\%. For larger models such as RankED~\cite{cetinkaya2024ranked}, DiffusionEdge~\cite{ye2024diffusionedge}, and SAUGE~\cite{liufu2024sauge}, the increase becomes negligible, less than 0.02\%.


\noindent{\textbf{Running time.}} We compare the runtime of \MethodLPP with standard post-processing on a CPU (Intel\textsuperscript{\textregistered} Xeon\textsuperscript{\textregistered} E5-2630 v3), as shown in Table \ref{tab:abl_exp}(d). Standard post-processing applies NMS once, followed by skeleton-based thinning for 100 thresholded edge maps ($[0,1)$ with a step of 0.01), resulting in 100 thinning operations per image during evaluation. For fairness, we implement vectorized PyTorch versions of NMS and thinning. \MethodLPP achieves lower runtime than NMS with a single thinning operation and is about 0.02\% of the cost of NMS with 100 thinning operations. Also, note that our method’s running time does not depend on the crispness of the given edge map, whereas both NMS and thinning are affected by it. 

\begin{table*}
\centering
\caption{\textbf{Exp 3 and 4:} Effect of (a) confidence threshold $\tau_c$, (b) distance threshold $\TD$, and (c) confidence score weight $\alpha$ on the BSDS dataset using PiDiNet. The baseline (PiDiNet) performance is ODS = .789 and OIS = .803. (d) CPU running time comparison of \MethodLPP and standard post-processing on the NYUD dataset using PiDiNet \cite{su2021pixel}. The reported running time corresponds to the entire NYUD test set (654 images).}

\resizebox{0.9\textwidth}{!}{%
\begin{tabular}{lccc@{\hskip 15pt}|lccc@{\hskip 15pt}|lccc @{\hskip 40pt}| l r}
\toprule
\multicolumn{4}{c}{\textbf{(a) Confidence threshold $\tau_c$}} &
\multicolumn{4}{c}{\textbf{(b) Distance threshold $\TD$}} &
\multicolumn{4}{c}{\textbf{(c) Confidence score weight $\alpha$}} &
\multicolumn{2}{c}{\textbf{(d) CPU Runtime on NYUD}} \\
\cmidrule(lr){1-4} \cmidrule(lr){5-8} \cmidrule(lr){9-12} \cmidrule(lr){13-14}
$\tau_c$ & \textbf{ODS} & \textbf{OIS} & \textbf{AP} &
$\TD$ & \textbf{ODS} & \textbf{OIS} & \textbf{AP} &
$\alpha$ & \textbf{ODS} & \textbf{OIS} & \textbf{AP} &
\textbf{Method} & \textbf{Time (sec.)} \\
\midrule
0.01 & \textbf{.800} & \textbf{.811} & \textbf{.866} &
2 & .797 & .807 & .856 &
5 & .630 & .639 & .653 &
NMS & 25.69 \\
0.03 & .799 & .811 & .858 &
4 & \textbf{.800} & \textbf{.811} & \textbf{.866} &
10 & .783 & .790 & .827 &
NMS + Thin. ($\times$1) & 41.36 \\
0.05 & .799 & .811 & .845 &
6 & .797 & .809 & \textbf{.866} &
15 & .794 & .802 & .845 &
NMS + Thin. ($\times$100) & 1875.57 \\
0.10 & .799 & .808 & .836 &
8 & .797 & .808 & \textbf{.866} &
20 & .798 & \textbf{.811} & .857 &
\MethodLPP~(Ours) & 32.98 \\
0.20 & .788 & .803 & .820 &
--- & --- & --- & --- &
25 & \textbf{.800} & \textbf{.811} & .866 &
--- & --- \\
0.30 & .761 & .771 & .803 &
--- & --- & --- & --- &
30 & .798 & \textbf{.811} & \textbf{.867} &
--- & --- \\
\bottomrule
\end{tabular}
}
\label{tab:abl_exp}
\end{table*}




\noindent\textbf{GPU memory overhead.} We analyze the additional GPU memory consumption introduced by \MethodLPP.
For $80 \times 80$ inputs, the overhead remains modest (1.72~GB) even under a low confidence threshold ($\TD=0.01$), where almost all pixels in the crisp edge map $\EdgeC$ are included in the matching process. In contrast, processing $320 \times 320$ inputs incurs significantly higher memory usage, reaching 28.32~GB at the same threshold. To mitigate this overhead, we divide $320 \times 320$ input into four non-overlapping patches and process them independently. This patch-wise strategy substantially reduces memory requirements, enabling the method to run on low-memory GPUs while leading to longer training times. Therefore, the input size can be adapted according to the available GPU memory. Moreover, this analysis is based on the first training iteration. As the model learns to predict sharper edge maps with fewer noisy responses, the number of pixels that exceed the confidence threshold $\TC$ decreases, leading to lower memory usage in later training stages. For detailed GPU memory overhead, see the supplementary material.

\subsection{Exp 5: Qualitative Results}

\begin{figure}
\centering\footnotesize
\includegraphics[width=1.\linewidth]{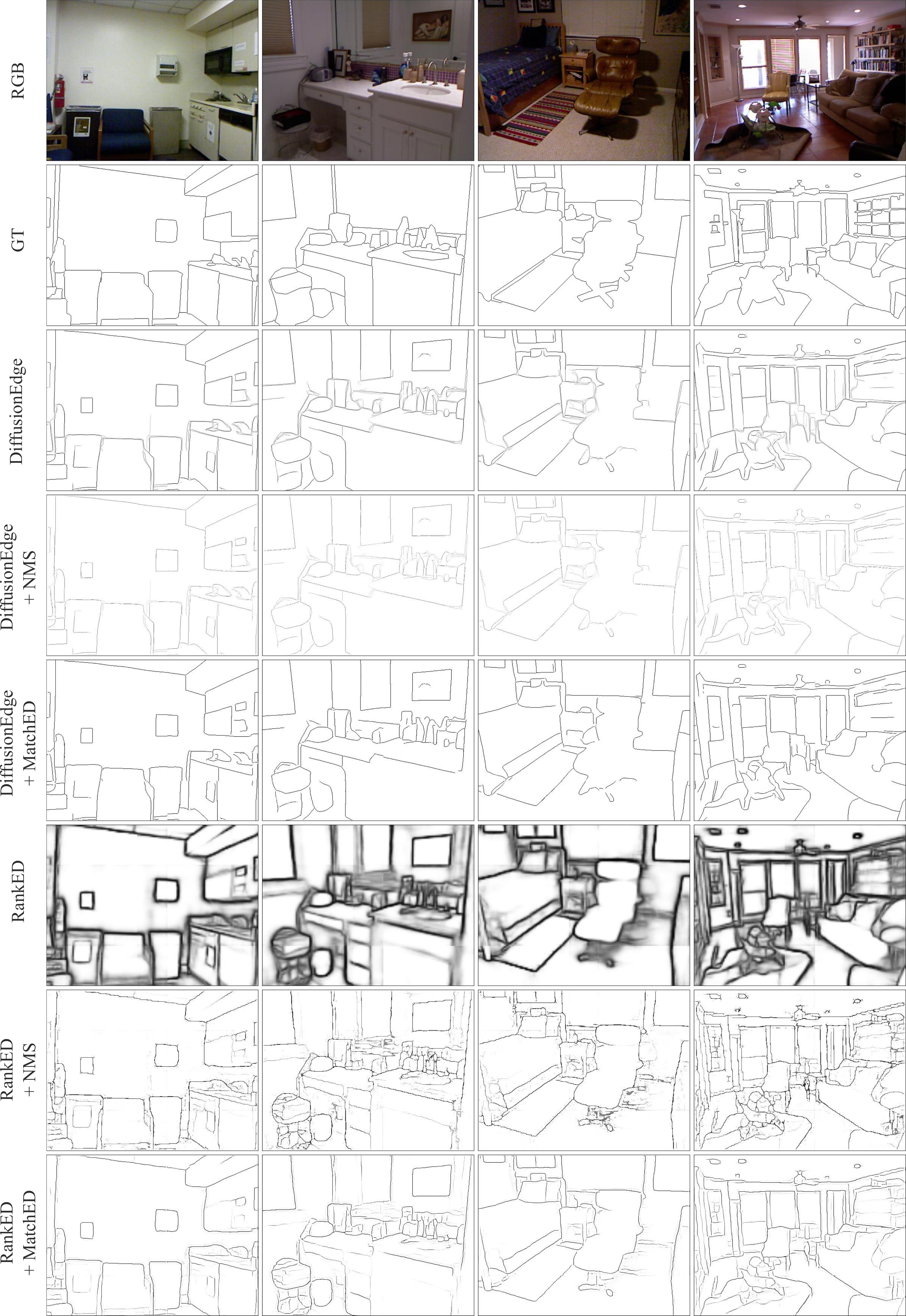}
  \caption{\textbf{Exp 5:} Visual results of the raw outputs of SOTA models, their results after applying NMS, and their \MethodLPP integrated versions on NYUD-v2 dataset. Best viewed by zooming in. Baseline results are generated from the pipeline of \MethodLPP.}
\label{fig:visRes}
\end{figure}

We present a comparison between the raw outputs of SOTA models, their results after applying NMS, and their \MethodLPP integrated versions within the \MethodLPP pipeline on the NYUD-v2 dataset in  \Cref{fig:visRes}. As shown in the figures, \MethodLPP successfully produces one-pixel-wide edge maps, remains robust to variations in edge thickness across different models and achieves better visual results than NMS. See the supplementary material for more visual results.




\section{Conclusion}
\label{sec:conc}
In this study, we propose a novel plug-and-play matching-based supervision for the crisp edge detection problem. Our method can be seamlessly integrated into the pipeline of any detection model, and allows for end-to-end training. Extensive experiments demonstrate that it is a robust solution, suitable for various architectures, including CNNs, transformers, and diffusion models, each with different loss functions. Notably, this is the first crisp edge detection model to match the performance of standard post-processing methods while also boosting performance in CEval metrics. Beyond metric improvements, it gives a competitive visual quality compared to traditional post-processing approaches.

\noindent\textbf{Limitations.} Although our method uses the same number of hyperparameters as standard post-processing, tuning them requires retraining, which increases computational cost. However, our experiments suggest that \MethodLPP is relatively robust against hyperparameter changes.

\FloatBarrier

\section*{Acknowledgments}
This work was supported by the Council of Higher Education Research Universities Support program through METU Scientific Research Projects  (``New Techniques in Visual Recognition’’, Project No. ADEP-312-2024-11485). We also gratefully acknowledge the computational resources provided by METU-ROMER, Center for Robotics and Artificial Intelligence, as well as TUBITAK ULAKBIM Truba.

\bibliographystyle{ieeenat_fullname}
\bibliography{refs}

\end{document}


\maketitle

\tableofcontents

\section{Reasons for Edge Thickness}
Both traditional and learning-based edge detection methods tend to produce thick raw edge predictions. This behavior arises from three primary and compounding factors: 
(i) annotation misalignment, 
(ii) merged supervision, and 
(iii) inherent image gradients.

\noindent\textbf{Annotation Misalignment:} 
Human annotators inevitably introduce small spatial inconsistencies when manually tracing edges in the scene. These pixel-level deviations prevent perfect alignment between RGB images and their ground-truths. As a result, during training, the supervision signal is spatially dispersed rather than concentrated on a single pixel. Consequently, models learn to activate not only the true boundary pixel but also its neighboring pixels, leading to thicker predictions.

\noindent\textbf{Merged Supervision:} 
Datasets such as BSDS \cite{arbelaez2010contour} and MultiCue \cite{mely2016systematic} provide multiple annotations per image. Annotators often disagree not only on precise boundary localization but also on the semantic validity of certain boundaries. When these differing annotations are aggregated into a single ground-truth map, the resulting supervision inherently broadens in spatial extent. Training with such thick supervision signals encourages models to produce similarly thick raw predictions.

\noindent\textbf{Inherent Image Gradients:} 
In natural RGB images, transitions between distinct regions rarely occur as ideal one-pixel step edges. Due to optical blur, illumination changes, sensor sampling, and other imaging factors, physical boundaries typically appear as gradual intensity transitions spanning multiple pixels. Therefore, even prior to annotation, the underlying image gradients already exhibit spatial thickness, further contributing to thick edge responses in learned models.

\MethodLPP explicitly addresses and mitigates all three factors, enabling the model to produce spatially precise and significantly thinner edge predictions.

\section{Time Complexity of Our Matching Algorithm}
\label{sup_sec:timeComp}
In this section, we provide a detailed analysis of the time complexity involved in generating the proposed matching-based supervision. First, computing the cost matrix $C$ in Eq. 3 of the main manuscript requires comparing all pixels in the ground-truth $G$ and the predicted crisp edge map $\EdgeC$. Therefore, the time complexity of this step is $\mathcal{O}\bigl((W \cdot H)^2\bigr)$, where $W$ and $H$ denote the width and height of $\EdgeC$. Then, solving the bipartite matching problem in Eq. 4 of the main manuscript using the linear sum assignment algorithm has a cubic time complexity of $\mathcal{O}\bigl((N_G \cdot N_{\EdgeC})^3\bigr)$.
Finally, recovering unmatched ground-truth edges in Eq. 6 of the main manuscript requires iterating over all pixels in the ground-truth $\GT$, and it has a time complexity of $\mathcal{O}\bigl((W \cdot H)\bigr)$.

Consequently, the total time complexity of the proposed matching-based supervision algorithm is
\begin{equation}
\mathcal{O}\bigl((W \cdot H)^2 + (N_G \cdot N_{\EdgeC})^3 + (W \cdot H)\bigr).
\end{equation}
The dominant term is the cubic cost of the linear sum assignment. In practice, $N_{\EdgeC}$ can be large in the initial iterations due to noisy and thick edge predictions. However, as the model learns a more accurate and crisp edge map, $N_{\EdgeC}$ decreases significantly, thereby reducing the computational cost in subsequent iterations.

\section{Details of Light-weight CNN}
\begin{figure}
\centering
\includegraphics[width=0.7\linewidth]{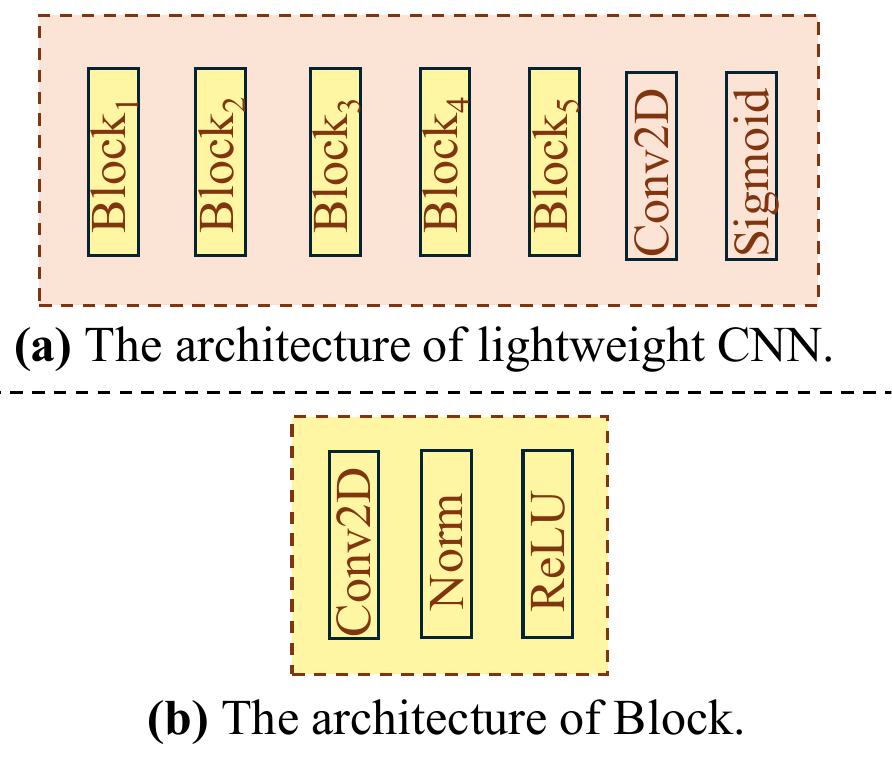}
  \caption{ (a) A lightweight CNN consists of five blocks followed by a final Conv2D layer with a sigmoid activation. Therefore, it has only approximately 21K parameters. (b) Each block contains Conv2D, ReLU, and normalization layers, where the type of normalization (e.g., BatchNorm, LayerNorm) is chosen to match that used in the base edge detector.} 
  \label{supp_fig:lightCNN}
\end{figure}

As shown in Figure \ref{supp_fig:lightCNN}(a), a light-weight CNN has five blocks and one Conv2D layer with sigmoid activation. Each block (Figure \ref{supp_fig:lightCNN}(b)) has one Conv2D layer with ReLU activation and one normalization layer. The normalization layer type matches that of the base edge detector. For example, PiDiNet \cite{su2021pixel} + \MethodLPP uses BatchNorm2D, whereas SAUGE \cite{liufu2024sauge} + \MethodLPP uses LayerNorm. In total, it introduces only $\sim$21k parameters. 

\section{Hyperparameter Configurations}
Hyperparameter settings for each base model and dataset are presented in Table \ref{supp_tab:hyper_param}. We tuned the $\beta$ coefficient, the loss weight of \MethodLPP in Eq. 8 of the main manuscript, only for RankED \cite{cetinkaya2024ranked}, due to its grid-like artifacts \cite{yang2024denoising} and down-sampled predictions. To mitigate the performance degradation caused by these two factors, we performed a dedicated search for $\beta$ only on RankED.

For tuning the parameters $\alpha$ (the weight of the confidence score) and $\TC$ (the confidence threshold used to select edge pixels) in Eq. 3 of the main manuscript, our heuristic is that less noisy and crisper raw edge predictions benefit from a larger $\TC$ (e.g., 0.1) and a smaller $\TC$ (e.g., 5).

\begin{table}[hbt!]
\caption{Hyperparameter configurations used for integrating \MethodLPP~into each base model across NYUD-v2 \cite{silberman2012indoor}, BSDS \cite{arbelaez2010contour}, Multi-Cue \cite{mely2016systematic}, and BIPED \cite{soria2023dense} datasets. $\TC$ is the confidence threshold for selecting edge pixels in Eq. 4 of the main manuscript, $\alpha$ is the weight of the confidence score in Eq. 3 of the main manuscript, and $\beta$ is the loss coefficient of \MethodLPP.}
\label{supp_tab:hyper_param}
\centering
\resizebox{0.8\columnwidth}{!}{%
\begin{tabular}{lcccc}
\toprule
Model & Dataset & $\TC (\times10^{-2})$ & $\alpha$ & $\beta$ \\
\midrule
\multirow{4}{*}{RankED \cite{cetinkaya2024ranked}} 
 & NYUD & 1 & 25 & 5 \\
 & BSDS & 1 & 25 & 5 \\
 & Multi-Cue & 1 & 25 & 5 \\
 & BIPED & -- & -- & -- \\
\midrule
\multirow{4}{*}{PiDiNet \cite{su2021pixel}} 
 & NYUD & 1 & 20 & 1 \\
 & BSDS & 1 & 25 & 1 \\
 & Multi-Cue & 1 & 55 & 1 \\
 & BIPED & 1 & 55 & 1 \\
\midrule
\multirow{4}{*}{DiffusionEdge \cite{ye2024diffusionedge}}
 & NYUD & 10 & 5 & 1 \\
 & BSDS & 10 & 5 & 1 \\
 & Multi-Cue & -- & -- & -- \\
 & BIPED & 10 & 5 & -- \\
\midrule
\multirow{4}{*}{SAUGE \cite{liufu2024sauge}}
 & NYUD & 5 & 20 & 1 \\
 & BSDS & 5 & 20 & 1 \\
 & Multi-Cue & -- & -- & -- \\
 & BIPED & -- & -- & -- \\
\bottomrule
\end{tabular}
}
\end{table}

\section{Results on Multi-cue}
\begin{table}[t]
\caption{Comparison of SOTA results on Multi-Cue dataset. While \MethodLPP~ results are reported under CEval, other results are reported under SEval. ODS: Optimal Dataset Score, OIS: Optimal Image Score, AP: Average Precision, AC: Average Crispness. Best and second-best results are shown in bold and underlined, respectively.}
\label{supp_tab:ch5_SOTA_crisp_Multicue}
\centering
\resizebox{1.\columnwidth}{!}{%
\begin{tabular}{llccc}
\toprule
\multicolumn{5}{c}{\textbf{Edge}} \\
\midrule
\multicolumn{1}{c}{\textbf{Method}} & \textbf{ODS} & \textbf{OIS} & \textbf{AP} \\
\midrule
Human & & .750$\pm$.024 & -- & -- \\
Multicue \cite{mely2016systematic}  \kaynak{(VR’16)} & .830$\pm$.002 & -- & -- \\
HED \cite{xie2015holistically}  \kaynak{(ICCV’15)} & .851$\pm$.014 & .864$\pm$.011 & -- \\
RCF \cite{liu2017richer}  \kaynak{(CVPR’17)} & .857$\pm$.004 & .862$\pm$.004 & -- \\
BDCN \cite{he2019bi}  \kaynak{(CVPR’19)} & .891$\pm$.001 & .898$\pm$.002 & .935$\pm$.002 \\
\highlightA{PiDiNet} \cite{su2021pixel}  \kaynak{(ICCV’21)} & .855$\pm$.007 & .860$\pm$.005 & -- \\
EDTER \cite{pu2022edter}  \kaynak{(CVPR’22)} & .894$\pm$.005 & .900$\pm$.003 & .944$\pm$.002 \\
UAED \cite{zhou2023treasure}  \kaynak{(CVPR’23)} & .895$\pm$.002 & .902$\pm$.001 & .949$\pm$.002 \\
MuGE \cite{zhou2024muge}  \kaynak{(CVPR’24)} & .898$\pm$.004 & .900$\pm$.004 & .950$\pm$.004 \\
Diff.Edge \cite{ye2024diffusionedge}  \kaynak{(AAAI’24)} & .904 & .909 & -- \\
\highlightB{RankED} \cite{cetinkaya2024ranked}  \kaynak{(CVPR’24)} & \textbf{.962$\pm$.003} & \textbf{.965$\pm$.003} & \textbf{.973$\pm$.006} \\
\midrule
\multicolumn{1}{l}{\highlightA{PiDiNet} $+$ \MethodLPP}& .903$\pm$.002 & .911$\pm$.004 & .973$\pm$.007 \\
\multicolumn{1}{l}{\highlightB{RankED} $+$ \MethodLPP} & \underline{.924$\pm$.010} & \underline{.929$\pm$.010} & \underline{.969$\pm$.007} \\
\midrule
\multicolumn{5}{c}{\textbf{Boundary}} \\
\midrule
\multicolumn{1}{c}{\textbf{Method}} & \textbf{ODS} & \textbf{OIS} & \textbf{AP} \\
\midrule
Human & & .760$\pm$.017 & -- & -- \\
Multicue \cite{mely2016systematic}  \kaynak{(VR’16)} & .720$\pm$.014 & -- & -- \\
HED \cite{xie2015holistically}  \kaynak{(ICCV’15)} & .814$\pm$.011 & .822$\pm$.008 & .869$\pm$.015 \\
RCF \cite{liu2017richer}  \kaynak{(CVPR’17)} & .817$\pm$.004 & .825$\pm$.005 & -- \\
BDCN \cite{he2019bi}  \kaynak{(CVPR’19)} & .836$\pm$.001 & .846$\pm$.003 & .893$\pm$.001 \\
\highlightA{PiDiNet} \cite{su2021pixel}  \kaynak{(ICCV’21)} & .818$\pm$.003 & .830$\pm$.005 & -- \\
EDTER \cite{pu2022edter}  \kaynak{(CVPR’22)} & .861$\pm$.003 & .870$\pm$.004 & .919$\pm$.003 \\
UAED \cite{zhou2023treasure}  \kaynak{(CVPR’23)} & .864$\pm$.004 & .872$\pm$.006 & .927$\pm$.006 \\
MuGE \cite{zhou2024muge}  \kaynak{(CVPR’24)} & .875$\pm$.006 & .879$\pm$.006 & \underline{.932$\pm$.006} \\
\highlightB{RankED} \cite{cetinkaya2024ranked}  \kaynak{(CVPR’24)} & \textbf{.963$\pm$.002} & \textbf{.967$\pm$.002} & \textbf{.995$\pm$.001} \\
\midrule
\multicolumn{1}{l}{\highlightA{PiDiNet} $+$ \MethodLPP} & .825$\pm$.003 & .829$\pm$.002 & .889$\pm$.003 \\
\multicolumn{1}{l}{\highlightB{RankED} $+$ \MethodLPP} & \underline{.941$\pm$.009} & \underline{.946$\pm$.008} & \textbf{.995$\pm$.006} \\

\bottomrule
\end{tabular}
}
\end{table}
Since Multi-Cue dataset lacks an official split, results are reported as the mean and variance over three random splits, following \cite{su2021pixel,zhou2024muge,cetinkaya2024ranked}, which results in longer training times compared to other datasets. Due to resource constraints, \MethodLPP is integrated into only two models, (i) PiDiNet \cite{su2021pixel} and (ii) RankED \cite{cetinkaya2024ranked}, and their performance is compared with SOTA in Table \ref{supp_tab:ch5_SOTA_crisp_Multicue}. Additionally, due to the lack of official models for Multi-Cue, SOTA methods are reported with post-processing, whereas our method is evaluated without post-processing.

When integrated into PiDiNet \cite{su2021pixel}, our method yields improvements of +4.8 and +5.1 in ODS and OIS, respectively, for the edge part, and scores of +0.7 and -0.1 for the boundary part. As discussed in Section 4.3 of the main manuscript, the down-scaled feature map size and interpolation artifacts cause the performance of \MethodLPP to drop to that of RankED \cite{cetinkaya2024ranked} when compared to the post-processed results.

\section{Effects of hyperparameter}

\begin{table*}[!htbp]
\centering
\caption{Effect of (a) confidence threshold $\tau_c$, (b) distance threshold $\TD$, and (c) confidence score weight $\alpha$ on BSDS dataset using PiDiNet. The baseline (PiDiNet) performance is ODS = .789 and OIS = .803.}

\resizebox{0.7\textwidth}{!}{%
\begin{tabular}{lccc@{\hskip 15pt}lccc@{\hskip 15pt}lccc}
\toprule
\multicolumn{4}{c}{\textbf{(a) Confidence threshold $\tau_c$}} &
\multicolumn{4}{c}{\textbf{(b) Distance threshold $\TD$}} &
\multicolumn{4}{c}{\textbf{(c) Confidence score weight $\alpha$}} \\
\cmidrule(lr){1-4} \cmidrule(lr){5-8} \cmidrule(lr){9-12}
\textbf{Param.} & \textbf{ODS} & \textbf{OIS} & \textbf{AP} &
\textbf{Param.} & \textbf{ODS} & \textbf{OIS} & \textbf{AP} &
\textbf{Param.} & \textbf{ODS} & \textbf{OIS} & \textbf{AP} \\
\midrule
0.01 & \textbf{.800} & \textbf{.811} & \textbf{.866} &
2 & .797 & .807 & .856 &
5 & .630 & .639 & .653 \\
0.03 & .799 & .811 & .858 &
4 & \textbf{.800} & \textbf{.811} & \textbf{.866} &
10 & .783 & .790 & .827 \\
0.05 & .799 & .811 & .845 &
6 & .797 & .809 & \textbf{.866} &
15 & .794 & .802 & .845 \\
0.10 & .799 & .808 & .836 &
8 & .797 & .808 & \textbf{.866} &
20 & .798 & \textbf{.811} & .857 \\
0.20 & .788 & .803 & .820 &
--- & --- & --- & --- &
25 & \textbf{.800} & \textbf{.811} & .866 \\
0.30 & .761 & .771 & .803 &
--- & --- & --- & --- &
30 & .798 & \textbf{.811} & \textbf{.867} \\
\bottomrule
\end{tabular}
}
\label{sup_tab:abl_exp}
\end{table*}

\noindent{\textbf{Effect of confidence threshold $\TC$.}} We investigate the effects of different confidence thresholds $\TC$ in Eq. 3 of the main manuscript on the performance of \MethodLPP, using PiDiNet \cite{su2021pixel} on BSDS dataset, as shown in Table \ref{sup_tab:abl_exp}(a). According to the results, the best performance is achieved when nearly all predictions (i.e., pixels with confidence > 0.01) are included in the matching process. Threshold values up to 0.1 yield similar performance, except for the AP metric. Excluding low-confidence pixels results in a lower AP score. Nevertheless, our method still outperforms standard post-processing for threshold values up to 0.2. During these experiments, the value of the confidence score $\alpha$ and distance threshold $\TD$ are used as 25 and 4, respectively.

\noindent{\textbf{Effect of distance threshold $\TD$.}} We analyze how varying the distance threshold $\TD$ in Eq. 3 of the main manuscript impacts the performance of \MethodLPP on BSDS dataset using PiDiNet \cite{su2021pixel}, as presented in Table \ref{sup_tab:abl_exp}(b). According to the results, all tested threshold values yield similar performance, except for $\TD = 2$ in the AP metric, which shows a 1.0-point lower score. Notably, the best performance is achieved with $\TD = 4$, which aligns with the distance threshold used during evaluation. During these experiments, the weight of the confidence score $\alpha$ and the confidence threshold $\TC$ are used as 25 and 0.01, respectively.

\noindent{\textbf{Effect of confidence score's weight $\alpha$.}} We examine the impact of varying the confidence score weight $\alpha$ in Eq. 3 of the main manuscript on the performance of \MethodLPP on BSDS dataset, when integrated into PiDiNet \cite{su2021pixel}, as shown in Table \ref{sup_tab:abl_exp}(c). The results show that $\alpha = 5$ leads to a significant performance drop across all metrics. The best performance is achieved with $\alpha = 25$, while $\alpha = 30$ yields nearly the same results. Moreover, using $\alpha = 15$ or $20$ still outperforms standard post-processing methods. During these experiments, the distance threshold $\TD$ and the confidence threshold $\TC$ are used as 2 and 0.01, respectively.

\noindent{\textbf{Heuristic for parameter choice.}} Our experiments indicate that tuning the distance threshold $\TD$ is unnecessary, as it has minimal impact on performance and works well when aligned with the evaluation threshold.

For the $\TC$ parameter, there is a trade-off between training time and performance: increasing $\TC$ reduces training time because fewer pixels are included in the matching process (see Section \ref{sup_sec:timeComp}), but slightly lowers performance. For lower computational cost, $\TC$ can be set in the range 0.1–0.2; for optimal performance, all pixels can be used, i.e., $\TC = 0.01$. For crisp and accurate raw edge predictions, such as \cite{ye2024diffusionedge}, setting $\TC = 0.1$ yields good performance.

The $\alpha$ parameter should be small for crisp raw results (e.g., $\alpha = 5$) and larger for noisy or thick raw results (e.g., $\alpha = 20$).
\section{GPU memory overhead}

Table \ref{tab:ch5_MemReq_comparison} presents the GPU memory overhead introduced by \MethodLPP under varying image resolutions and confidence thresholds $\TC$. As expected, memory consumption increases substantially with larger input sizes, rising from around 1.7 GB for $80\times80$ inputs to over 28 GB for $320\times320$ inputs. Moreover, increasing the confidence threshold $\TC$ leads to decreasing GPU memory usage, as a higher $\TC$ reduces the number of edge pixels involved in the matching process. Note that the memory consumption is sensitive to the number of edge pixels in the ground-truth as well as the number of pixels exceeding the confidence threshold $\TC$. As these numbers increase, the required GPU memory grows accordingly. 

\begin{table}
\centering
\caption{GPU memory overhead introduced by \MethodLPP at different image sizes and confidence thresholds $\TD$.}
\resizebox{0.75\columnwidth}{!}{%
\begin{tabular}{c c c}
\toprule
\textbf{Image Size} & \textbf{$\TD$} & \textbf{GPU Memory (GB)} \\
\midrule
\multirow{3}{*}{$80 \times 80$}  
  & 0.01 & 1.72 \\
  & 0.05 & 1.65 \\
  & 0.10 & 1.57 \\
\midrule
\multirow{3}{*}{$320 \times 320$} 
  & 0.01 & 28.43 \\
  & 0.05 & 27.30 \\
  & 0.10 & 25.86 \\
\bottomrule
\end{tabular}
}
\label{tab:ch5_MemReq_comparison}
\end{table}

\section{More Visual Results}
This section presents visual results on the NYUD, BSDS, BIPED, and Multi-Cue datasets. To enable a fair comparison for each base edge detector with an officially released checkpoint (on NYUD and on BSDS), we report two types of raw visual results:

\begin{itemize}
\item \textbf{Official Results}: Edge maps produced directly from the publicly released model weights. These serve as the canonical reference for each base model.
\item \textbf{\MethodLPP's Pipeline}: Raw edge maps produced by the same base model when used within the \MethodLPP's training pipeline. These results serve as the proper baseline for visually assessing how \MethodLPP refines the raw predictions.
\end{itemize}

Minor visual and numerical discrepancies between these two outputs are expected. Such differences arise from (i)  non-determinism in modern training pipelines (e.g., random initialization and data loading order), and (ii) the influence of \MethodLPP's loss gradients on the base model during an end-to-end optimization.

Across all datasets and base models, the visual results demonstrate that \MethodLPP consistently generates superior crisp edge maps. Notably, \MethodLPP successfully achieves thinning without altering the thickness of the raw input edge map.

\subsection{More Visual Results on NYUD}
\begin{figure*}[b]
\centering
\includegraphics[width=1\linewidth]{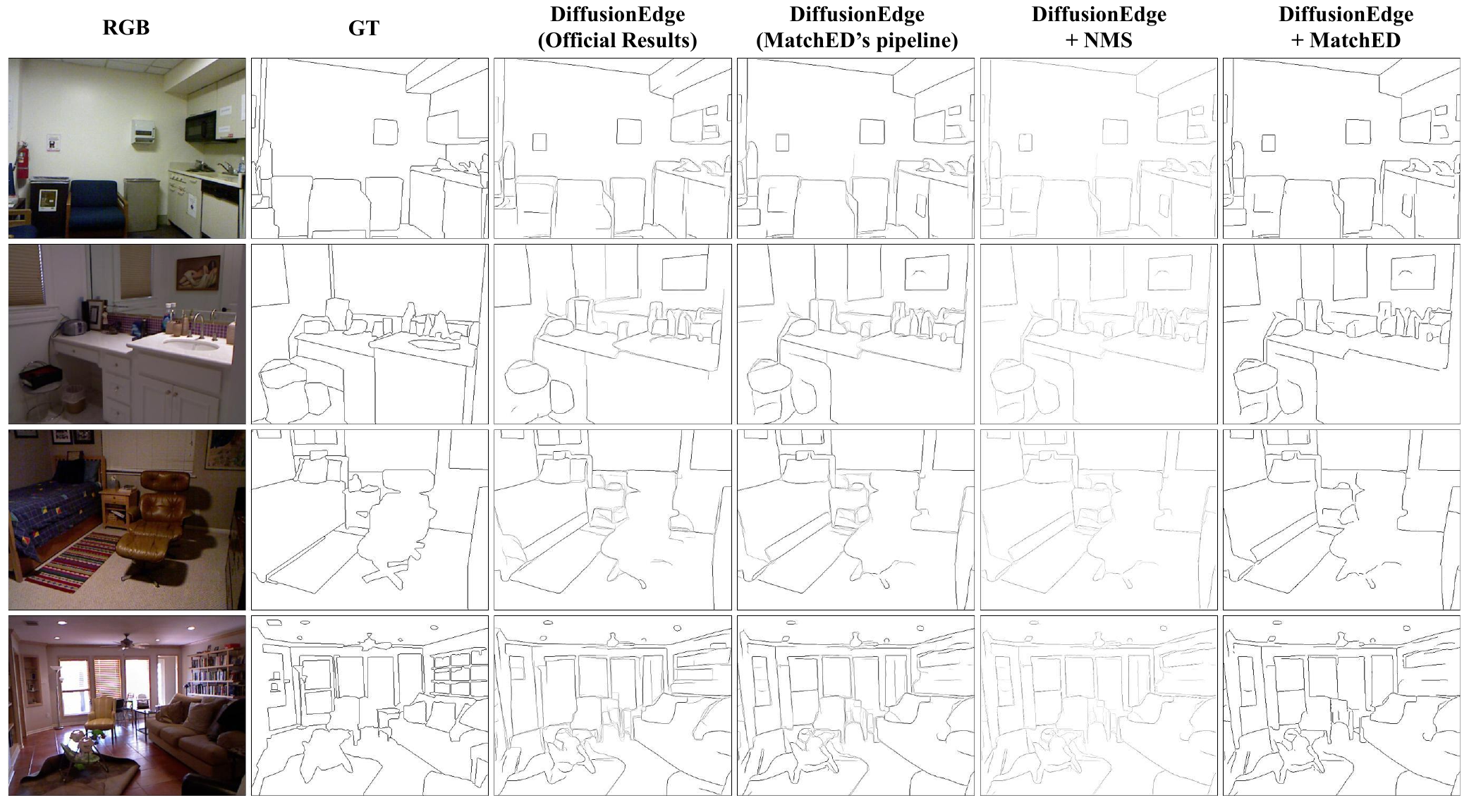}
  \caption{Qualitative comparisons on NYUD dataset using DiffusionEdge \cite{ye2024diffusionedge}. We show results from DiffusionEdge \cite{ye2024diffusionedge} (official checkpoint), raw outputs from our \MethodLPP pipeline, raw outputs after applying NMS, and their corresponding MATCHED integrated results, respectively. Best viewed zoomed-in.} 
  \label{supp_fig:nyud_diffusionEdge}
\end{figure*}

\begin{figure*}
\centering
\includegraphics[width=1\linewidth]{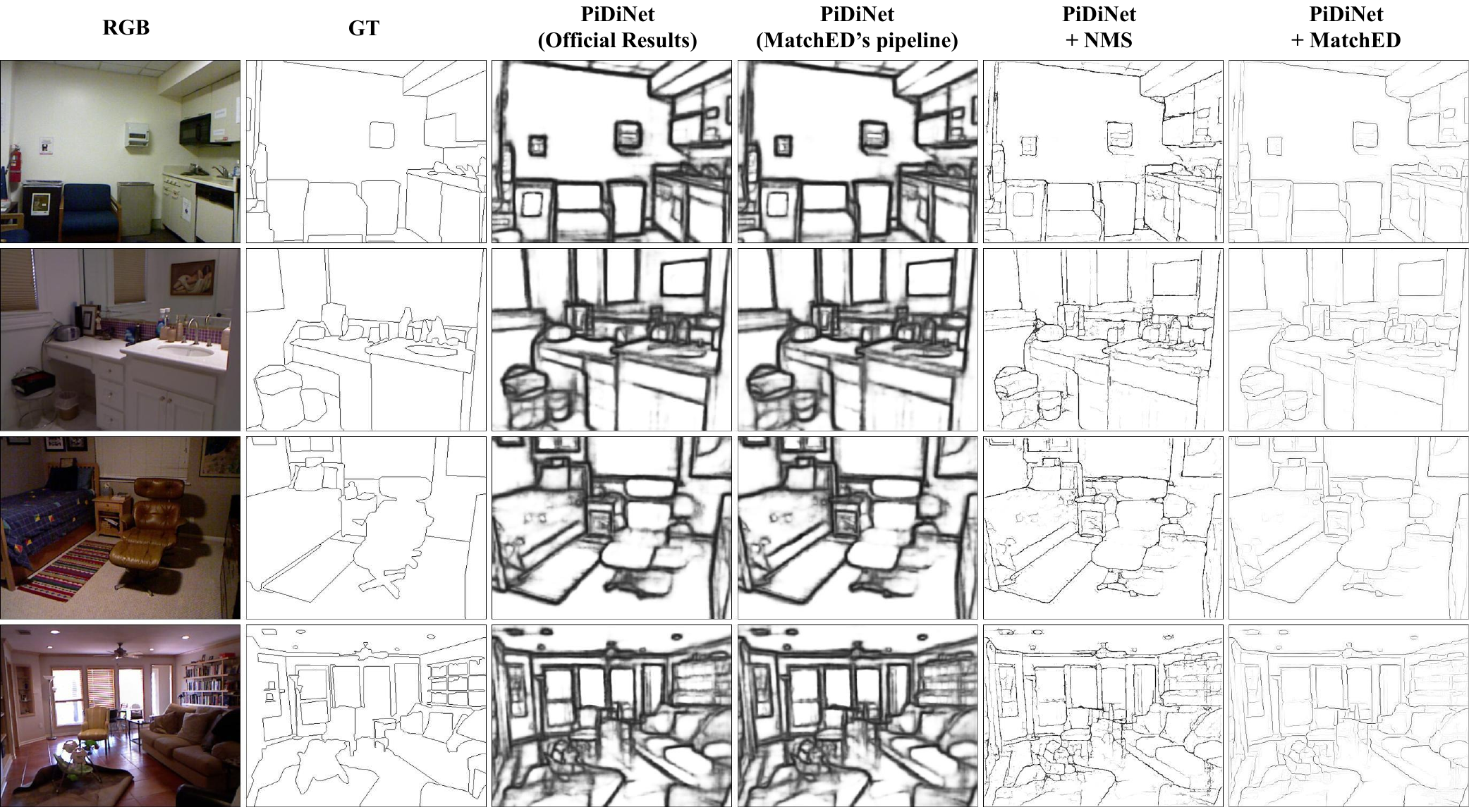}
  \caption{Qualitative comparisons on NYUD dataset using PiDiNet \cite{su2021pixel}. We show results from PiDiNet \cite{su2021pixel} (official checkpoint), raw outputs from our \MethodLPP pipeline, raw outputs after applying NMS, and their corresponding MATCHED integrated results, respectively. Best viewed zoomed-in.} 
  \label{supp_fig:nyud_pidinet}
\end{figure*}

\begin{figure*}
\centering
\includegraphics[width=1\linewidth]{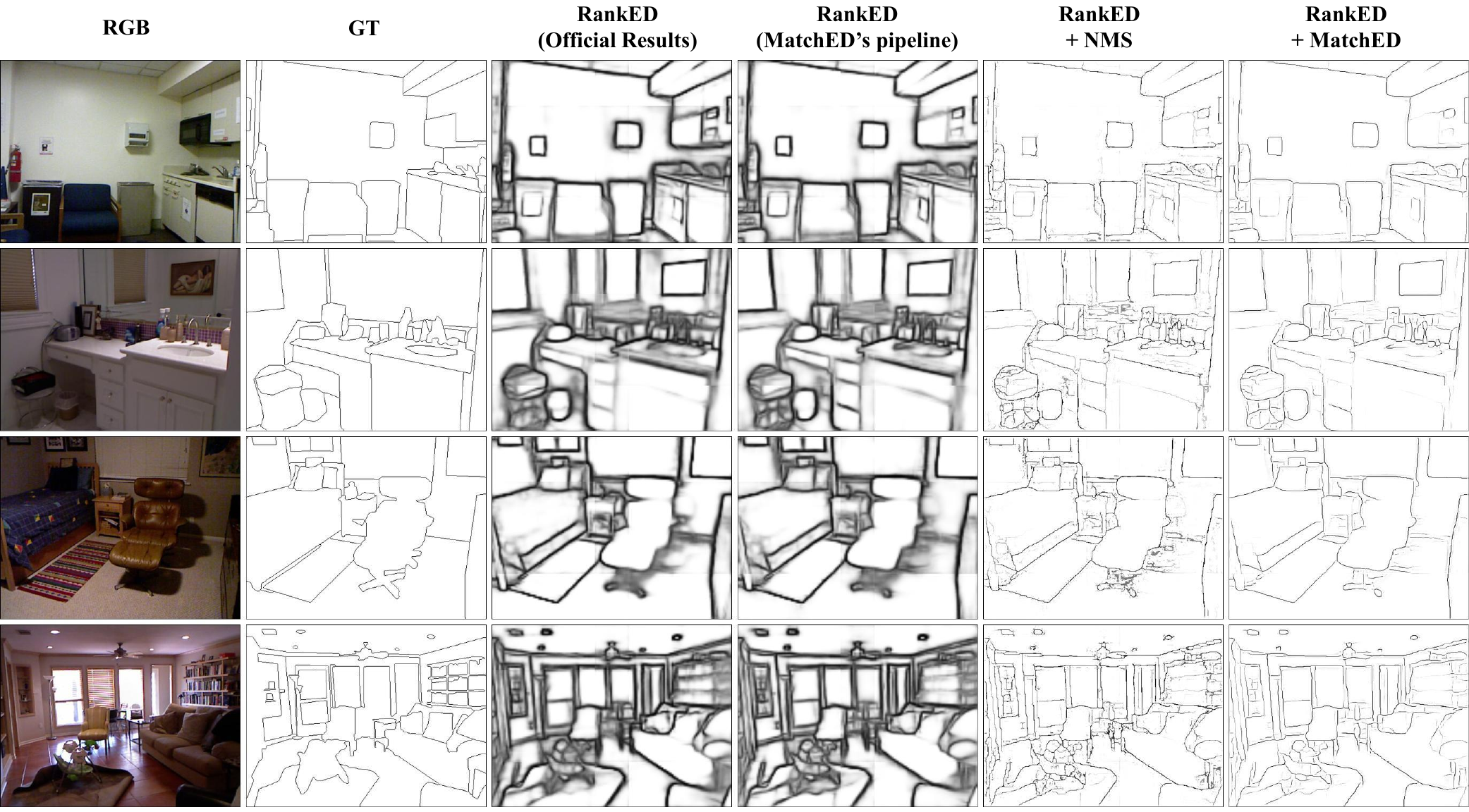}
  \caption{Qualitative comparisons on NYUD dataset using RankED \cite{cetinkaya2024ranked}. We show results from RankED \cite{cetinkaya2024ranked} (official checkpoint), raw outputs from our \MethodLPP pipeline, raw outputs after applying NMS, and their corresponding MATCHED integrated results, respectively. Best viewed zoomed-in.} 
  \label{supp_fig:nyud_ranked}
\end{figure*}

\begin{figure*}
\centering
\includegraphics[width=1\linewidth]{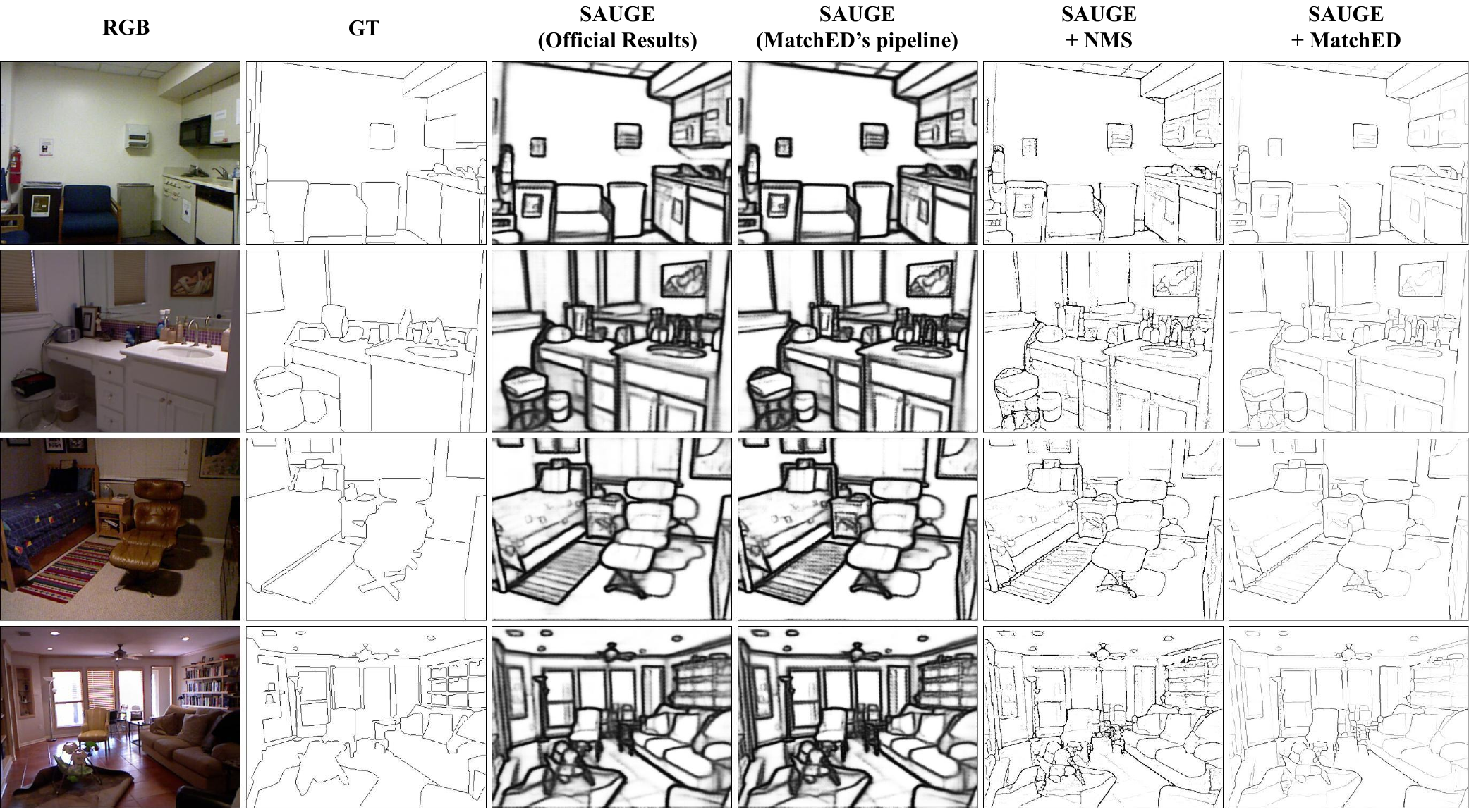}
  \caption{Qualitative comparisons on NYUD dataset using SAUGE \cite{liufu2024sauge}. We show results from SAUGE \cite{liufu2024sauge} (official checkpoint), raw outputs from our \MethodLPP pipeline, raw outputs after applying NMS, and their corresponding MATCHED integrated results, respectively. Best viewed zoomed-in.} 
  \label{supp_fig:nyud_sauge}
\end{figure*}

In this section, we present visual results of \MethodLPP on NYUD-v2 \cite{silberman2011indoor} dataset using the four base models: (i) DiffusionEdge \cite{ye2024diffusionedge} (Figure \ref{supp_fig:nyud_diffusionEdge}), (ii) PiDiNet \cite{su2021pixel} (Figure \ref{supp_fig:nyud_pidinet}), (iii) RankED \cite{cetinkaya2024ranked} (Figure \ref{supp_fig:nyud_ranked}), and SAUGE \cite{liufu2024sauge} (Figure \ref{supp_fig:nyud_sauge}). Each figure compares the raw outputs from official model checkpoints, the outputs from the \MethodLPP pipeline before and after NMS, and the outputs of \MethodLPP integrated versions.

\subsection{More Visual Results on BSDS}
\begin{figure*}
\centering
\includegraphics[width=1\linewidth]{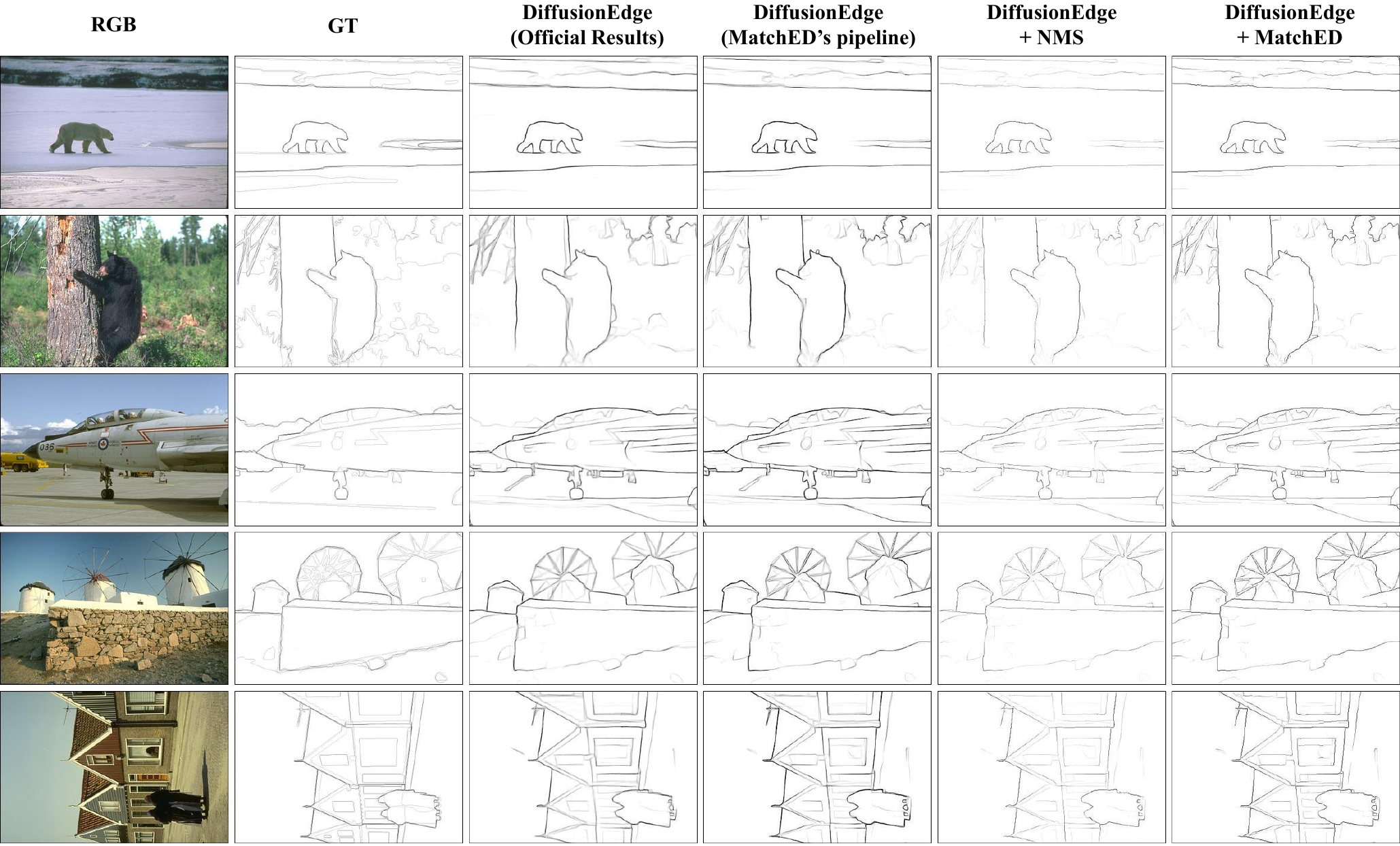}
  \caption{Qualitative comparisons on BSDS dataset using DiffusionEdge \cite{ye2024diffusionedge}. We show results from DiffusionEdge \cite{ye2024diffusionedge} (official checkpoint), raw outputs from our \MethodLPP pipeline, raw outputs after applying NMS, and their corresponding MATCHED integrated results, respectively. Best viewed zoomed-in.} 
  \label{supp_fig:bsds_diffusionEdge}
\end{figure*}

\begin{figure*}
\centering
\includegraphics[width=1\linewidth]{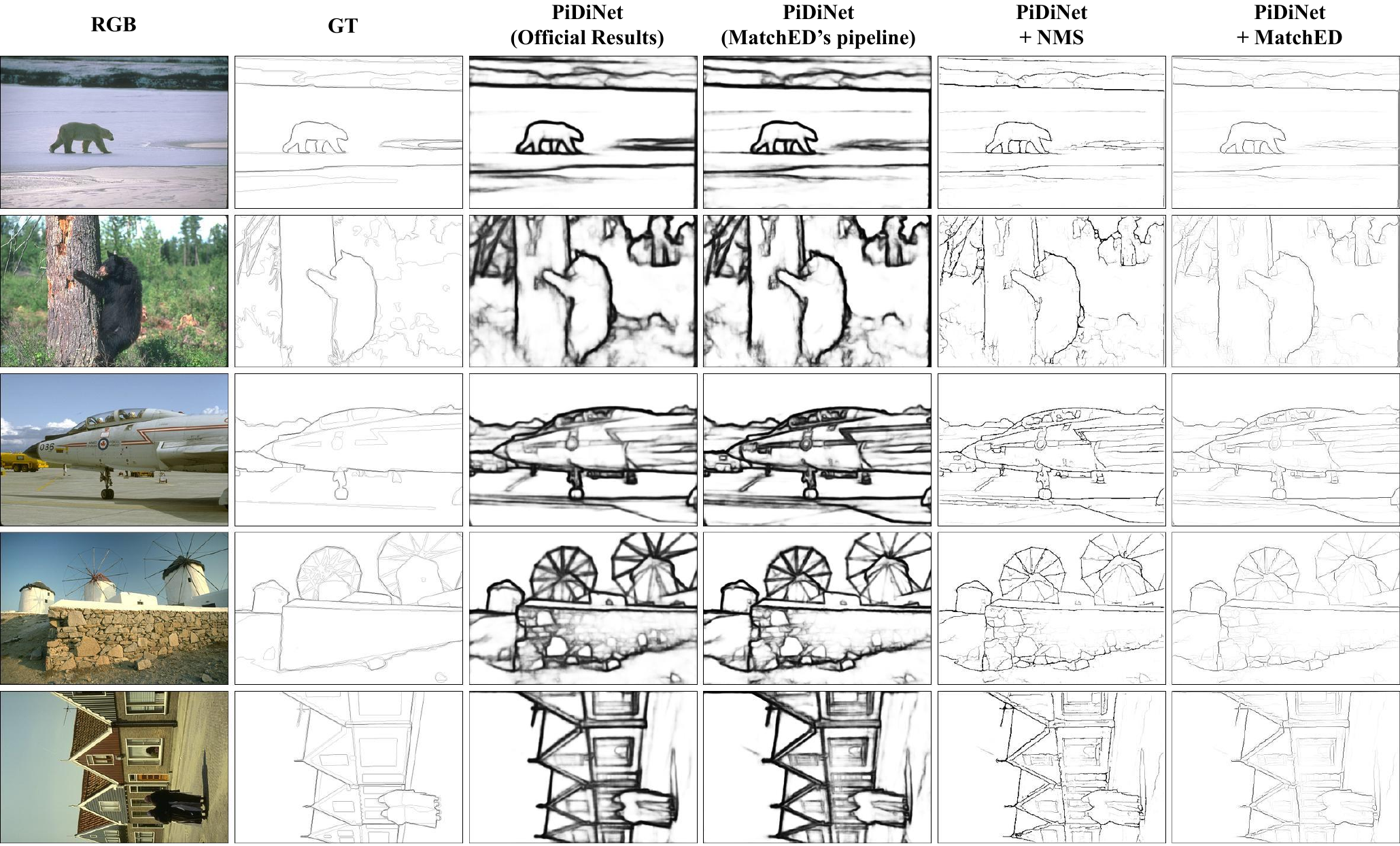}
  \caption{Qualitative comparisons on BSDS dataset using PiDiNet \cite{su2021pixel}. We show results from PiDiNet \cite{su2021pixel} (official checkpoint), raw outputs from our \MethodLPP pipeline, raw outputs after applying NMS, and their corresponding MATCHED integrated results, respectively. Best viewed zoomed-in.} 
  \label{supp_fig:bsds_pidinet}
\end{figure*}

\begin{figure*}
\centering
\includegraphics[width=1\linewidth]{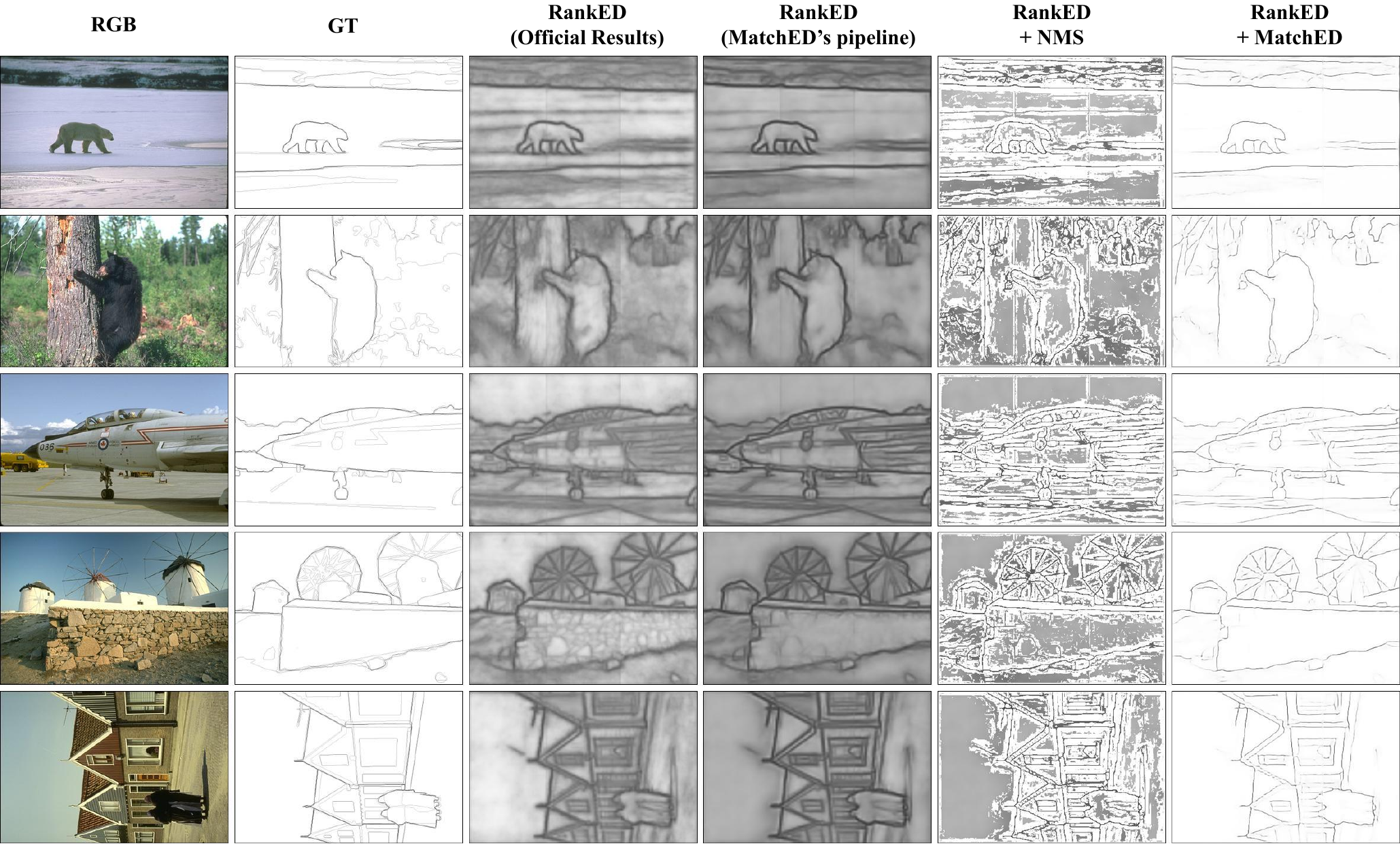}
  \caption{Qualitative comparisons on BSDS dataset using RankED \cite{cetinkaya2024ranked}. We show results from RankED \cite{cetinkaya2024ranked} (official checkpoint), raw outputs from our \MethodLPP pipeline, raw outputs after applying NMS, and their corresponding MATCHED integrated results, respectively. Best viewed zoomed-in.} 
  \label{supp_fig:bsds_ranked}
\end{figure*}

\begin{figure*}
\centering
\includegraphics[width=1\linewidth]{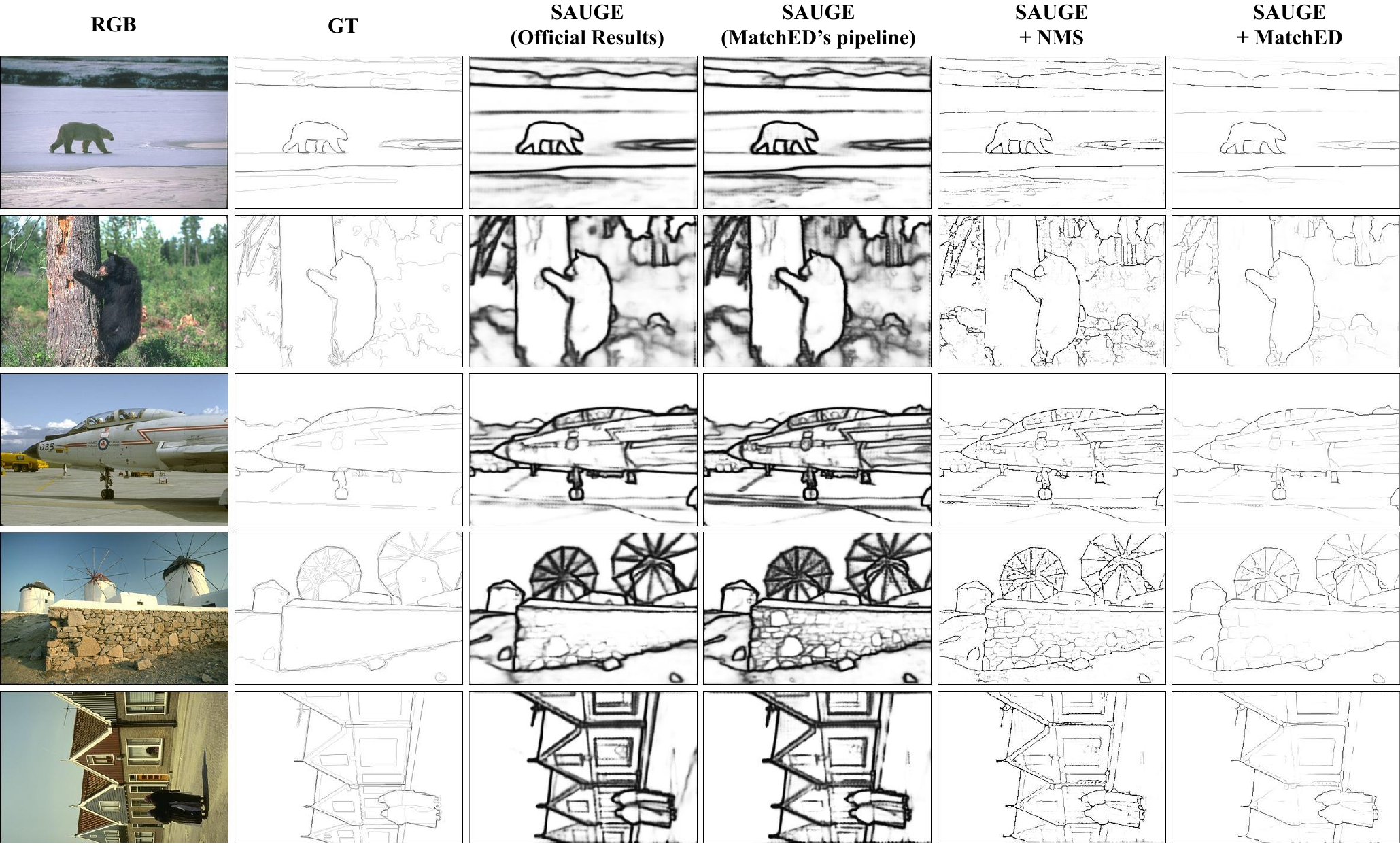}
  \caption{Qualitative comparisons on BSDS dataset using SAUGE \cite{liufu2024sauge}. We show results from SAUGE \cite{liufu2024sauge} (official checkpoint), raw outputs from our \MethodLPP pipeline, raw outputs after applying NMS, and their corresponding MATCHED integrated results, respectively. Best viewed zoomed-in.} 
  \label{supp_fig:bsds_sauge}
\end{figure*}

In this section, we present visual results of \MethodLPP on BSDS \cite{arbelaez2010contour} dataset using the four base models: (i) DiffusionEdge \cite{ye2024diffusionedge} (Figure \ref{supp_fig:bsds_diffusionEdge}), (ii) PiDiNet \cite{su2021pixel} (Figure \ref{supp_fig:bsds_pidinet}), (iii) RankED \cite{cetinkaya2024ranked} (Figure \ref{supp_fig:bsds_ranked}), and SAUGE \cite{liufu2024sauge} (Figure \ref{supp_fig:bsds_sauge}). Each figure compares the raw outputs from official model checkpoints, the outputs from the \MethodLPP pipeline before and after NMS, and the outputs of \MethodLPP integrated versions.

\subsection{More Visual Results on BIPED}
\begin{figure*}
\centering
\includegraphics[width=1\linewidth]{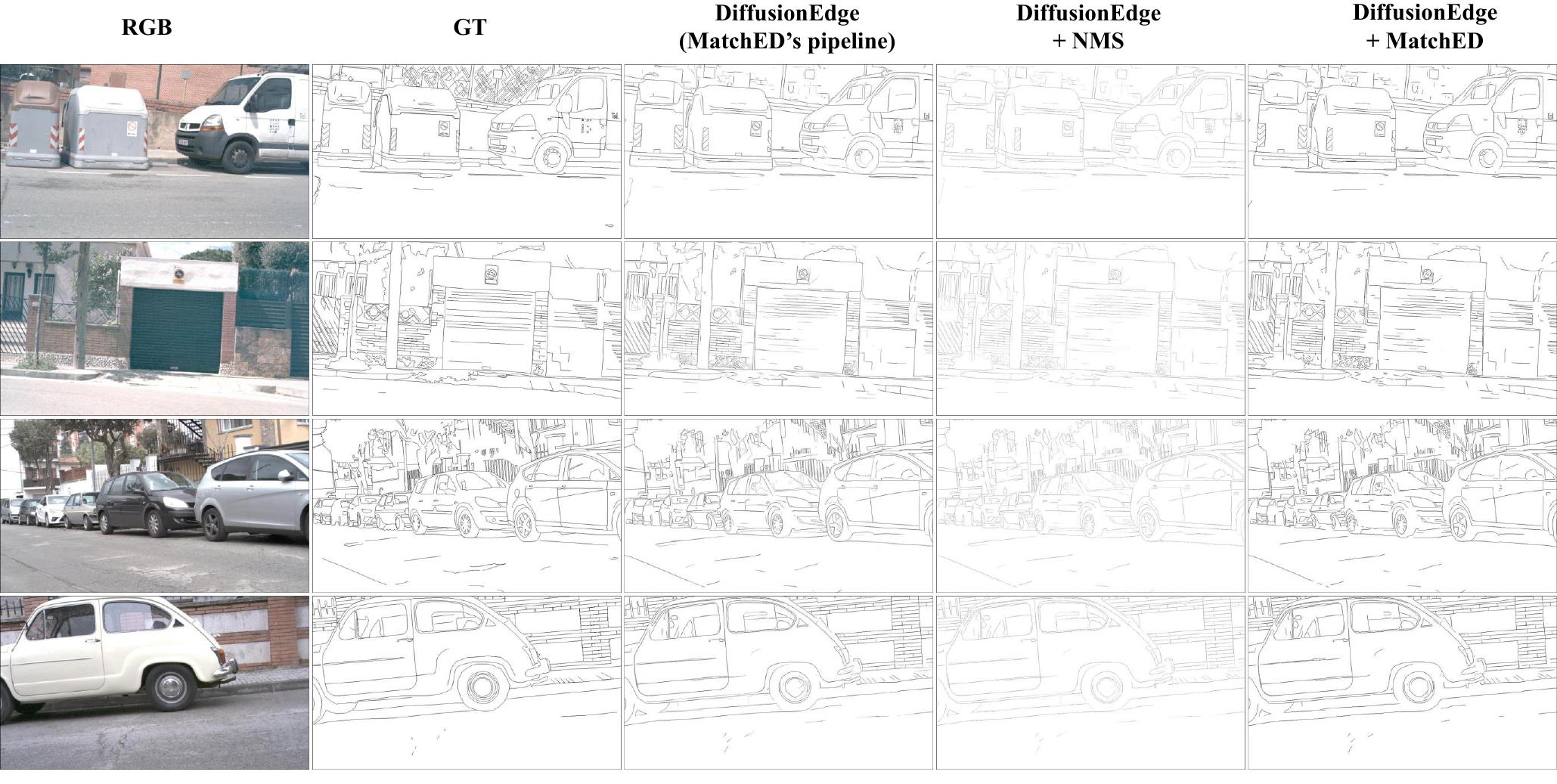}
  \caption{Qualitative comparisons on BIPED dataset using DiffusionEdge \cite{ye2024diffusionedge}. We show results of raw outputs from our \MethodLPP pipeline, raw outputs after applying NMS, and their corresponding MATCHED integrated version, respectively. Best viewed zoomed-in.} 
  \label{supp_fig:biped_diffusionEdge}
\end{figure*}

\begin{figure*}
\centering
\includegraphics[width=1\linewidth]{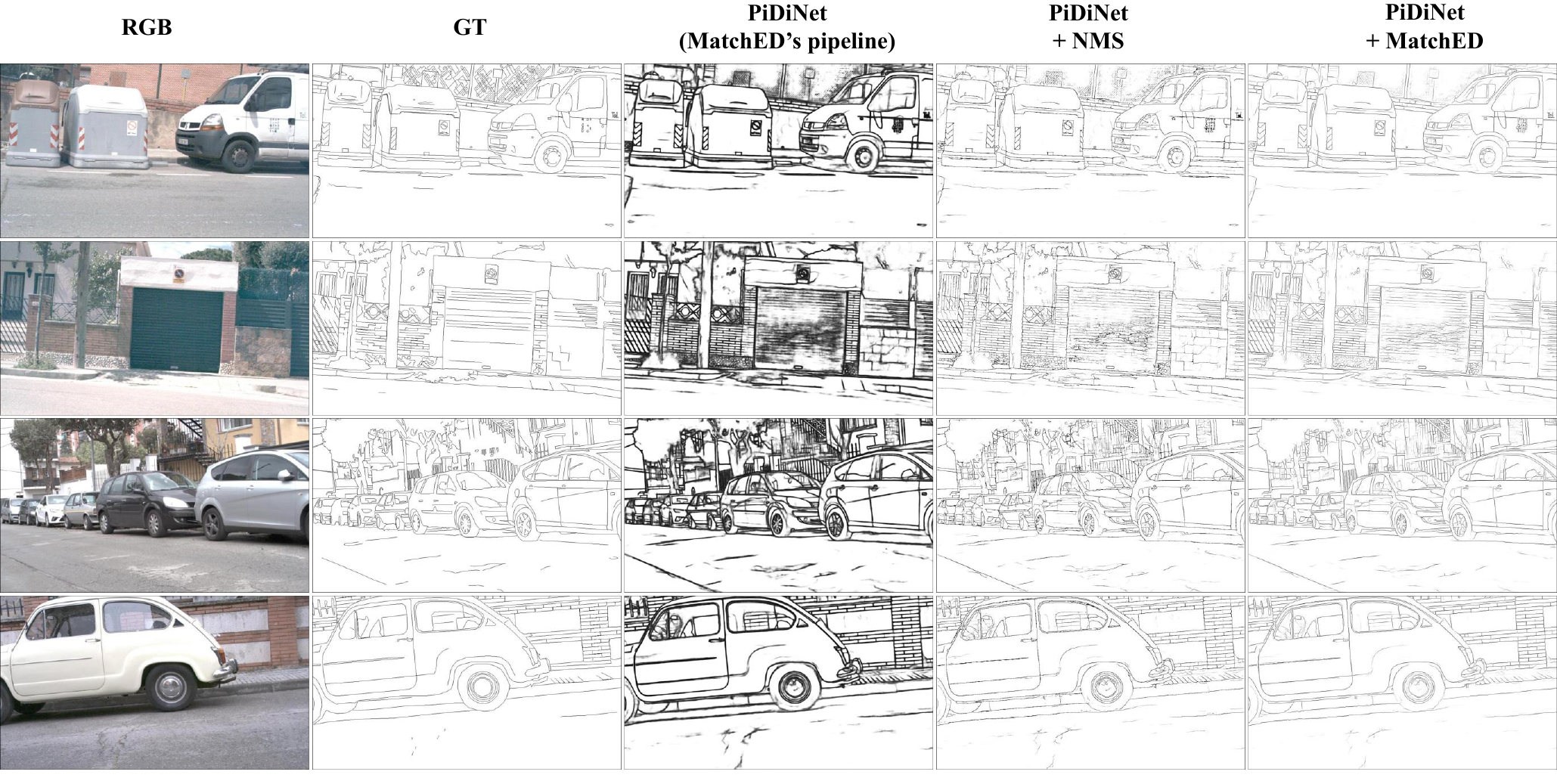}
  \caption{Qualitative comparisons on BIPED dataset using PiDiNet \cite{su2021pixel}. We show results of raw outputs from our \MethodLPP pipeline, raw outputs after applying NMS, and their corresponding MATCHED integrated version, respectively. Best viewed zoomed-in.} 
  \label{supp_fig:biped_pidinet}
\end{figure*}

In this section, we present visual results of \MethodLPP on BIPED-v2 \cite{soria2023dense} dataset using the two base models: (i) DiffusionEdge \cite{ye2024diffusionedge} (Figure \ref{supp_fig:biped_diffusionEdge}) and (ii) PiDiNet \cite{su2021pixel} (Figure \ref{supp_fig:biped_pidinet}).
Each figure compares the raw outputs from the \MethodLPP pipeline before and after NMS, and the outputs of \MethodLPP integrated versions.

\subsection{More Visual Results on Multi-Cue}
\begin{figure*}
\centering
\includegraphics[width=1\linewidth]{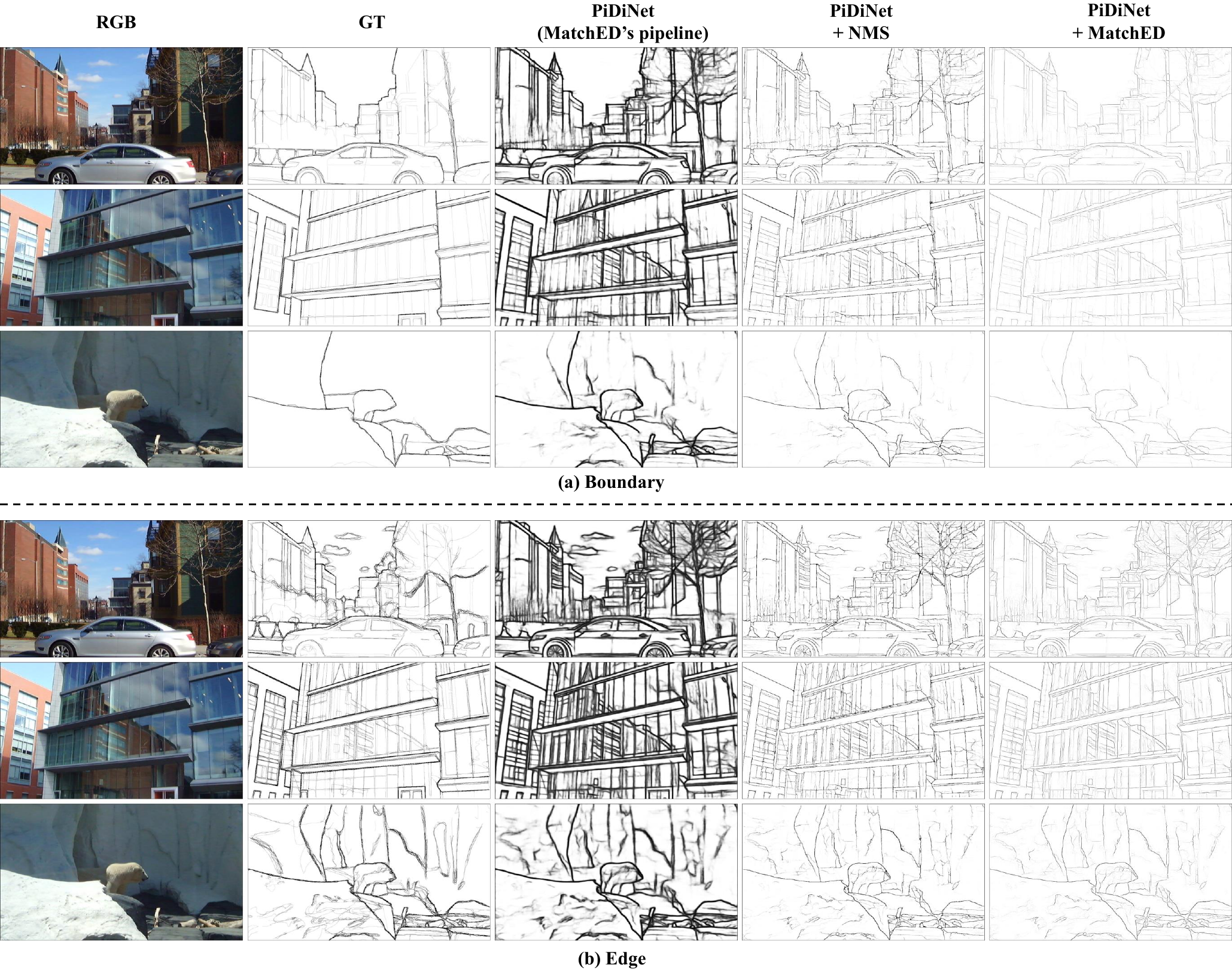}
  \caption{Qualitative comparisons on Multi-Cue dataset using PiDiNet \cite{su2021pixel}. We show results of raw outputs from our \MethodLPP pipeline, raw outputs after applying NMS, and their corresponding MATCHED integrated version, respectively. Best viewed zoomed-in. While the upper part shows the boundary detection results, the lower part shows the edge detection results.} 
  \label{supp_fig:mcue_pidinet}
\end{figure*}

\begin{figure*}
\centering
\includegraphics[width=1\linewidth]{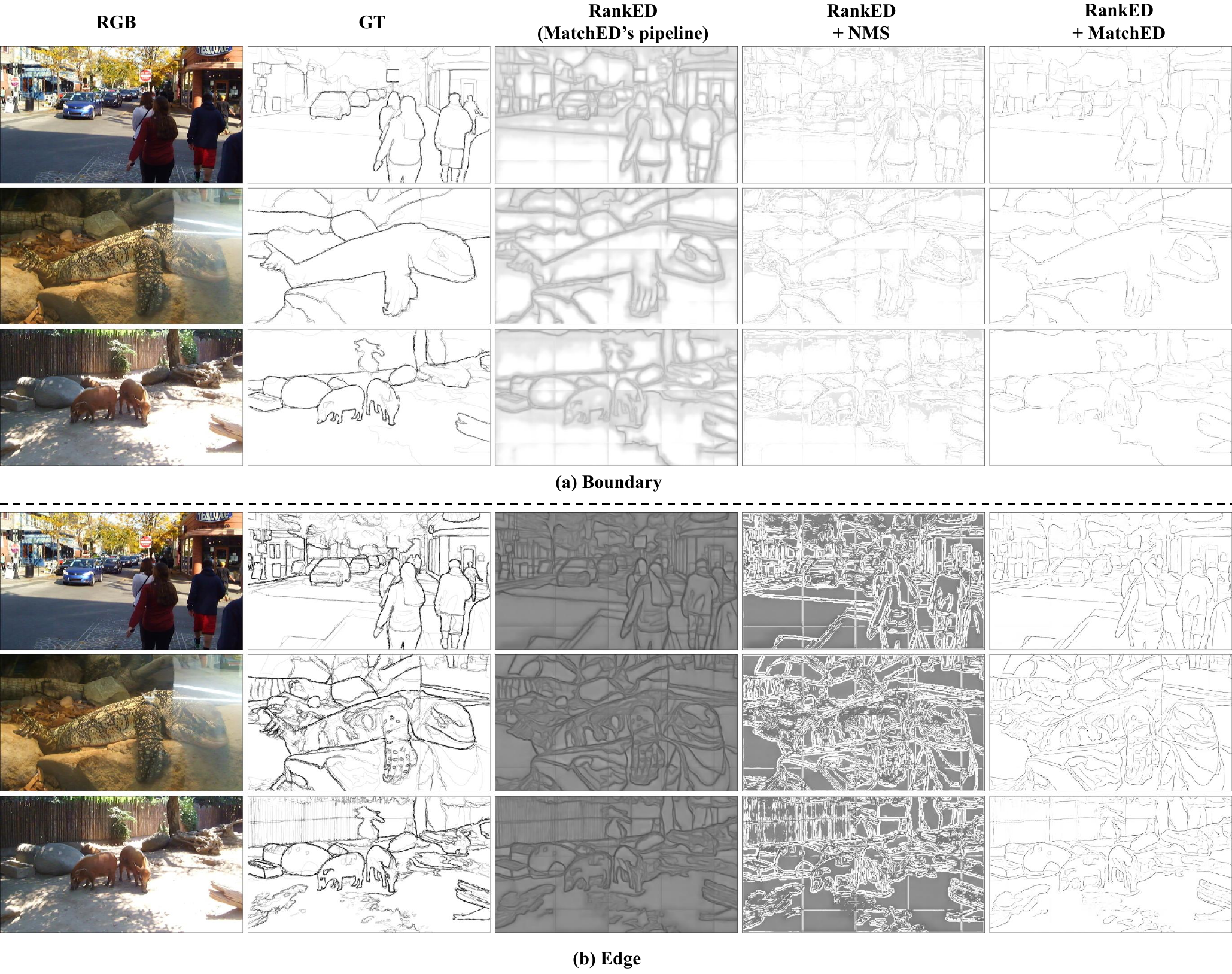}
  \caption{Qualitative comparisons on Multi-Cue dataset using RankED \cite{cetinkaya2024ranked}. We show results of raw outputs from our \MethodLPP pipeline, raw outputs after applying NMS, and their corresponding MATCHED integrated version, respectively. Best viewed zoomed-in. While the upper part shows the boundary detection results, the lower part shows the edge detection results.} 
  \label{supp_fig:mcue_ranked}
\end{figure*}

In this section, we present visual results of \MethodLPP on Multi-Cue \cite{mely2016systematic} dataset using the two base models: (i) PiDiNet \cite{su2021pixel} (Figure \ref{supp_fig:mcue_pidinet}) and (ii) RankED \cite{cetinkaya2024ranked} \ref{supp_fig:mcue_ranked}).

\FloatBarrier
\bibliographystyle{ieeenat_fullname}
\bibliography{refs}